\documentclass{article}
\usepackage[utf8]{inputenc}
\usepackage{amssymb}
\usepackage{graphicx}
\usepackage{tabularx}

\usepackage[final]{corl_2026} % Uncomment for the camera-ready ``final'' version.

\usepackage{color}
\usepackage{booktabs}
\usepackage[table]{xcolor}
\usepackage{tabularx}
\usepackage{array}
\usepackage{makecell}
\usepackage{pifont}
\usepackage{amsmath}
\usepackage{amsfonts}
\usepackage{longtable}
\usepackage{dsfont}
\usepackage{caption}
\usepackage{adjustbox}
\usepackage{hyperref}
\usepackage{tikz} 
\usepackage{graphicx} 
\usepackage{subfigure}

\usepackage{xcolor}  % for experiment legend
\definecolor{cRARM}{HTML}{1f77b4}
\definecolor{cRoboCLIP}{HTML}{d62728}
\definecolor{cTemporalOT}{HTML}{2ca02c}
\definecolor{cGVL}{HTML}{9467bd}
\definecolor{cRoboMeter}{HTML}{e377c2}
\definecolor{cRoboDopamine}{HTML}{ff7f0e}
\definecolor{cAblationSim}{HTML}{9f6559}
\newcommand{\leg}[2]{\textcolor{#1}{\rule[0.4ex]{0.4cm}{1.4pt}}\,\,#2}

\newcommand{\panel}[3]{%
  \begin{tabular}[t]{@{}c@{}}
    \parbox[c][2\baselineskip][c]{#1}{\scriptsize\centering #2} \\
    #3
  \end{tabular}%
}

\usepackage{wrapfig}   % provides wraptable
\usepackage{multirow}

\newcommand{\indicator}[1]{\mathds{1}\!\left\{#1\right\}}

\title{RARM: Confidence-Gated Progress Reward Modeling for RL in Manipulation}

\author{
\normalfont
Pengzhi Yang\,\textsuperscript{1,*},
Xinyu Wang\,\textsuperscript{2,*},
Pengyu Jing\,\textsuperscript{1,*},
Kehan Wen\,\textsuperscript{1,*},
Yiduo Qu\,\textsuperscript{1,3}, \\
Zhenhao Huang\,\textsuperscript{1},
Minghao Fu\,\textsuperscript{4},
Xin Liu\,\textsuperscript{1},
Yaheng Shen\,\textsuperscript{1},
Fan Shi\,\textsuperscript{1,$\sharp$}\\
\normalfont\textsuperscript{1}Human-Centered Robotic Lab, National University of Singapore \qquad
\normalfont\textsuperscript{2}Booking.com \\
\textsuperscript{3}University of Cambridge \qquad
\normalfont\textsuperscript{4}School of Artificial Intelligence, Nanjing University\\
\normalfont\textsuperscript{*}Equal contribution
\qquad
\textsuperscript{$\sharp$}Corresponding author
}

\begin{document}
\maketitle

%===============================================================================

\begin{abstract}

Reinforcement learning for robot manipulation is often bottlenecked by reward design, especially in long-horizon tasks: sparse success rewards provide weak supervision, while hand-crafted dense rewards are tedious to design and generalize poorly across tasks. Progress-based reward models offer a promising alternative by estimating how far an observation has advanced toward task completion, but existing approaches often require task-specific demonstrations or progress labels, and can assign high rewards to visually plausible but physically incorrect states. We introduce the Reference-Anchored Reward Model (RARM), a lightweight visual comparator that converts a single successful demonstration into a dense, progress-aware reward. RARM is trained once on general-purpose videos with a contrastive temporal objective, requiring no robot-specific data, task-specific reward labels, or per-task reward engineering. At deployment, RARM matches rollout clips to reference clips and rewards only confident forward progress, suppressing uncertain matches that may otherwise produce false-positive rewards. Across 9 simulated manipulation tasks from LIBERO and MetaWorld and 4 real-world tasks, RARM achieves the best overall success rates in subsequent RL training, with particularly large gains on long-horizon tasks such as cloth folding, where unreliable progress estimates are especially harmful. Our project webpage: \href{https://rarm-robotics.github.io/}{https://rarm-robotics.github.io/}.

\end{abstract}


%===============================================================================

\section{Introduction}

Imitation learning (IL) underlies many recent vision–language–action (VLA) models~\cite{team2024octo, zitkovich2023rt, chi2025diffusion, black2024pi_0, intelligence2025pi}, but policies trained by pure imitation generalize poorly to out-of-distribution states, where small errors compound and drive the robot into regions rarely covered by demonstrations~\cite{ross2011reduction, de2019causal, codevilla2019exploring}. Reinforcement learning (RL) offers a complementary route: through interaction, an agent can improve beyond demonstrated trajectories and acquire corrective and recovery behaviors. This is especially valuable in long-horizon and contact-rich manipulation tasks, but its central obstacle is reward design. Sparse outcome rewards give weak, delayed signal, while hand-crafted dense rewards are labor-intensive and struggle to generalize across tasks and embodiments. This motivates learned reward models that offer dense feedback automatically, most naturally by estimating task progress—how far the current observation has advanced toward completion.

Progress, however, is usable only when the model knows how much to trust it. A rollout can look near-complete in a coarse visual sense while being physically wrong and such false-positive progress prediction will be exploited by the RL agent~\citep{tian2026position, ayalew2025progressor, leng2025taming}. A practical progress reward should therefore be both data-light,  as opposed to the heavy demonstration budgets imitation learning requires—and confidence-aware, able to withhold reward when its own estimate is unreliable. 
\begin{wrapfigure}{r}{0.35\linewidth}
    \centering
    \includegraphics[
        width=\linewidth,
        trim={0 0 0 0},
        clip
    ]{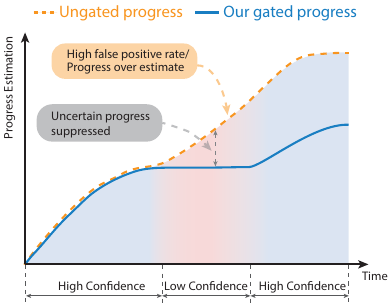}
    \caption{\textbf{Motivation.} Existing reward models often overestimate failed states, provide unreliable progress signals during failures. RARM addresses these failure modes with confidence-gate by rewarding only confident forward progress along the reference and suppressing uncertain matches.}
    \label{fig:motivation}
    \vspace{-1.0em}
\end{wrapfigure}
Existing methods rarely achieve both. VLM-based reward models~\cite{ma2024vision, lee2026roboreward, liang2026robometer, zhang2025rewind} 
carry broad semantic priors but are costly to query and over-score visually plausible failures; smaller task-specific models~\cite{chen2025sarm, liu2025timerewarder, ayalew2025progressor, yang2024rank2reward} cut inference cost but demand substantial per-task data and still overestimate progress under distribution shift. Approaches that explicitly model uncertainty, such as VIPER, Diffusion Reward, and PROGRESSOR~\cite{escontrela2023video, huang2024diffusion, ayalew2025progressor}, recognize that dense progress should not be trusted blindly—but they still rely on task-specific or task-related data, limiting their ability to transfer directly to new tasks.

We propose \textbf{Reference-Anchored Reward Model (RARM)}, a lightweight reward model that resolves this tension by turning a single successful demonstration into a confidence-gated dense progress reward. The demonstration acts as a progress anchor—a sequence for localizing how far a rollout has advanced, not a trajectory to imitate frame by frame. RARM compares each rollout clip against the reference clips to estimate progress, but never rewards raw similarity. Instead it rewards confident forward progress and suppresses uncertain matches, so misleading dense labels are withheld exactly when progress localization is unreliable. Crucially, this confidence signal arises naturally from the similarity comparison itself, needing no separate uncertainty model, failure data, or subtask labels. RARM is trained once on ordinary videos with a contrastive temporal objective; at RL time it needs only one reference demonstration per task and no expensive reward-model queries in the control loop.

Our contributions are summarized as follows:
\begin{itemize}
\item We introduce \textbf{RARM}, which turns a single successful demonstration into a dense progress reward with no task-specific reward supervision—achieving data-light, confidence-aware progress estimation where prior work meets only one of these criteria.
\item RARM rewards confident forward progress and explicitly filters out low-confidence matches, suppressing the false-positive rewards that drive reward hacking.
\item Across 9 simulated tasks (LIBERO~\citep{liu2023libero}, MetaWorld~\citep{yu2020meta}) and 4 real-world tasks, RARM achieves the best overall success rates with low-overhead reward queries, with especially large gains on long-horizon settings such as cloth folding.
\end{itemize}

\section{Related Works}
\label{sec:related works}

\subsection{Reward Model}

Learned reward models aim to replace brittle task-specific reward engineering with dense feedback derived from videos, demonstrations, language, or foundation-model priors. A major direction learns progress- or value-like rewards from visual and multimodal supervision, including value-/representation-based rewards~\cite{ma2022vip,ma2023liv}, temporal progress learning~\cite{liu2025timerewarder,ayalew2025progressor,yang2024rank2reward}, task-structured rewards for long-horizon manipulation~\cite{zhang2024universal,kim2025subtask,chen2025sarm}, and language- or VLM-conditioned reward models and critics~\cite{zhang2025rewind,yang2024adapt2reward,jain2025smooth,ma2024vision,zhai2025vision,lee2026roboreward,liang2026robometer,tan2025robo,chen2026topreward}. Another line grounds rewards in reference behavior through inverse reward learning, trajectory alignment, object/keypoint matching, or visual similarity~\cite{kumar2023graph,fu2024robot,shipoints2reward,guzey2025bridging,yu2025genflowrl}. Generative approaches instead use video prediction or diffusion models to score expert-likeness or task consistency~\cite{escontrela2023video,huang2024diffusion}, while automated reward construction from foundation models provides another route to dense feedback~\cite{ma2023eureka,ghasemipour2025self}.

Despite their effectiveness, these methods remain imperfect reward proxies. VLM-based rewards can be computationally expensive and prone to false positives on visually plausible but physically incorrect states; task-specific progress models often require in-domain videos, success/failure data, online refinement, or subtask annotations; and trajectory-, point-, or object-motion matching rewards can be sensitive to temporal misalignment, occlusion, and deviations from the demonstrated path. Generative uncertainty or push-back refinement can reduce overconfident rewards~\cite{huang2024diffusion,escontrela2023video,ayalew2025progressor}, but typically depends on task-related videos or online rollout data. More broadly, learned rewards may introduce dataset biases, shortcut correlations, overconfidence, or exploitable failure modes during RL~\cite{gao2023scaling,shen2023loose,tian2026position,leng2025taming}.

\subsection{Reinforcement Learning for Manipulation}

RL for manipulation has long been constrained by reward design: the sparse, long-horizon nature of manipulation forces task-specific shaping to navigate the exploration–exploitation trade-off~\citep{riedmiller2018learning}. A range of approaches has emerged to mitigate this limitation. Sim-to-real RL exploits privileged information in simulation for reward engineering and transfers via domain randomization~\citep{tobin2017domain, andrychowicz2020learning}; large-scale real-world RL bypasses simulation through massive on-robot data collection~\citep{kalashnikov2018scalable}; and sample-efficient real-world RL incorporates demonstrations and human interventions to reduce interaction cost~\citep{rajeswaran2017learning, luo2024serl}. Inverse RL and adversarial imitation approaches~\citep{ho2016generative, finn2016guided} instead infer rewards from expert trajectories, while hindsight relabeling~\citep{andrychowicz2017hindsight} and intrinsic motivation~\citep{pathak2017curiosity} alleviate exploration inefficiency without explicit shaping. Our work, in contrast, defines rewards via a general-purpose pixel-level visual comparator that bypasses task-specific reward engineering and requires only a single reference demonstration, in contrast to the large datasets typically needed by imitation-based methods~\citep{chi2025diffusion, zhao2023learning}.

\begin{figure}[t]
	\centering
	\includegraphics[
        width=\linewidth,
        trim={0 0 0 0},
        clip
    ]{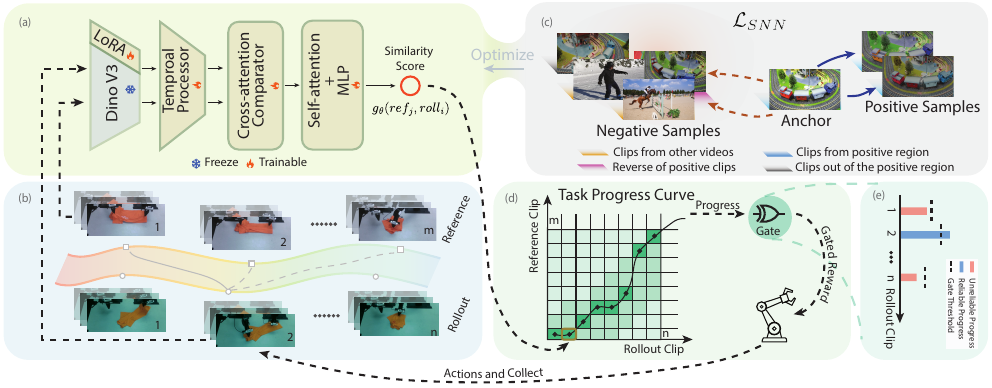}
	\caption{\textbf{Method Overview.} {Reward Model Training:} As shown in (c), we sample an anchor clip from an unlabeled video, positives from the same temporal region, and negatives from three complementary sources. These clip pairs are scored by the comparator in (a), which is trained with the soft-nearest-neighbours loss in Eq.~\eqref{eq:snnl}. {RL Training:} Given a reference video and a rollout video in (b), we compare each rollout clip with all reference clips to form the comparison matrix in (d). 
    % Progress is estimated from the best-matched reference clips and passed through a confidence gate to produce the final reward. The policy is then updated with this reward, and new rollouts are collected iteratively. The confidence threshold in (e) is computed automatically from self-comparisons within the reference video. 
    Rather than using these similarities directly as rewards, RARM uses them to estimate where the rollout lies along the reference, reading progress from the best-matched reference clips. To avoid false-positive progress, this estimate is passed through a confidence gate in (e): uncertain matches are suppressed, so the policy is not rewarded when progress localization is unreliable. The policy is updated with the resulting reward as new rollouts are collected iteratively. The confidence threshold is computed once from self-comparisons within the reference demonstration, which sets how strong a valid match should be.
    }
    \vspace{-0.5cm}
	\label{fig:overview}
\end{figure}

\section{Approach}
\label{sec:approach}

We introduce the \emph{Reference-Anchored Reward Model} (RARM), a lightweight video-clip comparator that derives confidence-gated progress rewards from a single reference demonstration. RARM is trained by self-supervised contrastive learning, requiring no action or progress labels, task annotations, or large-scale robot data. At deployment, it uses the reference demonstration as a progress anchor: rather than directly rewarding visual similarity, it estimates how far the agent's recent behavior has advanced along the reference and rewards only confident forward progress. Figure~\ref{fig:overview} provides an overview, and the following sections describe contrastive clip learning (Sec.~\ref{sec:approach:contrastive}), the model architecture (Sec.~\ref{sec:approach:arch}), and reward computation for reinforcement learning (Sec.~\ref{sec:approach:usage}).

\subsection{Contrastive Learning from Unlabeled Video}
\label{sec:approach:contrastive}

RARM compares short, fixed-length clips ($T{=}5$ frames). Its training only requires videos with directed temporal change: over the short horizon of a clip, the scene should evolve rather than remain static or immediately return to a previous state. This mild assumption holds for many natural 
% and robot-manipulation 
videos, allowing reward-model training without curated task demonstrations or reward labels.

\textbf{Anchors, positives, and negatives.} From each video, we select a contiguous \emph{anchor region} and sample an \emph{anchor} clip from it. \emph{Positives} are additional clips randomly drawn from the same anchor region; they share the anchor's local temporal context and should be close to it in the learned space. \emph{Negatives} come from three sources: (1) clips from disjoint regions of the same video, (2) time-reversed clips from the anchor region, which preserve appearance but reverse dynamics, and (3) clips from other videos, as seen in Figure 2(c).
% in the mini-batch, which depict different scenes and provide additional negatives at no extra cost. 

For data augmentation, we apply random resized cropping and color jitter independently to each clip, using the same augmentation parameters for all frames within that clip to preserve temporal consistency. All clips sampled from the same video share the same random horizontal flip. These augmentations encourage robustness to framing, illumination, color shifts, and viewpoint changes while preserving the temporal structure needed for clip comparison.

\textbf{Objective.} Let $g_\theta(\cdot,\cdot)$ denote the scalar clip similarity score produced by the comparator (Sec.~\ref{sec:approach:arch}). For an anchor $a$ with positive set $\mathcal{P}$ and negative set $\mathcal{N}$, we minimize the soft-nearest-neighbours loss~\citep{salakhutdinov2007learning}: \begin{equation} \mathcal{L}_{\mathrm{SNN}} = -\log \frac{\sum_{p\in\mathcal{P}}\exp\!\big(g_\theta(a,p)/\tau\big)} {\sum_{p\in\mathcal{P}}\exp\!\big(g_\theta(a,p)/\tau\big) +\sum_{n\in\mathcal{N}}\exp\!\big(g_\theta(a,n)/\tau\big)}, \label{eq:snnl} \end{equation} 
averaged over the batch, with temperature $\tau{=}0.1$. 
% Because each anchor region yields multiple valid positives, the numerator aggregates all positives rather than selecting a single pair.

\subsection{Model Architecture}
\label{sec:approach:arch}

The comparator $g_\theta$ maps a pair of clips to a scalar score, as illustrated in Figure~\ref{fig:overview}(a).

% \textbf{Frame encoder.} Each frame is embedded by a pretrained DINOv3~\citep{simeoni2025dinov3} with trainable LoRA adapters~\citep{hu2022lora} in its attention blocks. A $384{\times}384$ frame yields a $24{\times}24$ grid of $768$-d patch tokens, giving $T{\times}576$ tokens per clip. 

% \textbf{Temporal processor.} To make attention tractable, patch tokens are spatially average-pooled, reducing the number of tokens per frame. We add temporal rotary position encodings~\citep{su2024roformer} along the frame axis and apply a single multi-head self-attention layer over the resulting $T{\times}64$ tokens, producing a temporally aware clip representation. The frame encoder and temporal processor are shared across both clips during comparison, forming a Siamese branch.

% \textbf{Frame encoder.} Each frame is embedded by a pretrained DINOv3 encoder~\citep{simeoni2025dinov3} with lightweight LoRA adapters~\citep{hu2022lora} in its attention blocks, providing strong visual features.

% \textbf{Temporal processor.} To make temporal reasoning tractable, patch tokens are spatially pooled and augmented with temporal rotary position encodings~\citep{su2024roformer}. A single multi-head self-attention layer then mixes tokens across frames, producing a temporally aware clip representation. The frame encoder and temporal processor are shared across both clips during comparison, forming a Siamese branch.

\textbf{Clip encoder.} Given a short video clip, we first embed each frame with a pretrained DINOv3~\citep{simeoni2025dinov3} equipped with lightweight LoRA adapters~\citep{hu2022lora}. The resulting patch-token features are spatially pooled, augmented with temporal rotary position encodings~\citep{su2024roformer}, and processed by a lightweight self-attention layer~\citep{vaswani2017attention} across frames to produce a temporally aware clip representation. The clip encoder is shared across both inputs during comparison, forming a Siamese branch.

\textbf{Cross-attention comparator and head.} Given two clip representations $X$ and $Y$ from the clip encoder, we use $X$ tokens as queries and $Y$ tokens as keys and values in a cross-attention block, followed by a projection to a lower dimension. This produces a representation of $X$ conditioned on $Y$, making the comparison directional: ``how does the rollout clip compare to the reference clip?'' A lightweight head maps the comparison tokens to the scalar score $g_\theta(X,Y)$, where larger values indicate stronger temporal compatibility between the two motion clips.

\subsection{Reward Computation for Policy Learning}
\label{sec:approach:usage}

Using raw comparator scores $g_\theta$ directly as rewards can be unstable: the scores may jitter from step to step and destabilise training. We therefore use the comparator only to \emph{locate} progress along the reference, and derive reward from confident progress advancement.

Given a target task, we slide a fixed window over a single reference demonstration to obtain overlapping clips $\{r_j\}_{j=1}^{M}$. Each reference clip is assigned a progress value $\pi_j\in[0,1]$ according to its normalized temporal position in the demonstration; this requires no annotation and increases monotonically along the reference. During RL training, we form the current rollout clip $c_t$ from the agent's recent frames and score it against all reference clips: \begin{equation} j^\star(t)=\arg\max_{j} g_\theta(c_t,r_j). \end{equation} This all-to-all rollout--reference comparison is not tied to synchronized timestamps: the current clip can match any point along the reference, allowing the policy to move at a different speed, pause, or recover before advancing. A per-clip confidence threshold $\delta_j$, seen in Figure 2(e), is calibrated offline from the reference's own match-score statistics and gates whether the match is trustworthy. If the match is confident, the corresponding reference progress provides the current progress estimate; otherwise, progress is held at its previous value. This reflects the behavior we observe during RL: failures and out-of-distribution states often appear not as reliable backward progress, but as a loss of confident match to the reference. The progress estimate and reward are therefore:
% \begin{equation} 
% p_t = \begin{cases} \pi_{j^\star(t)}, & g_\theta\big(c_t,\,r_{j^\star(t)}\big)\ge\delta_{j^\star(t)},\\[4pt] p_{t-1}, & \text{otherwise,} \end{cases} \qquad r^{\mathrm{prog}}_t = \indicator{p_t > p_{t-1}} . \label{eq:reward} 
% \end{equation} 

\begin{equation}
p_t =
\begin{cases}
\pi_{j^\star(t)}, & g_\theta\big(c_t,\,r_{j^\star(t)}\big)\ge\delta_{j^\star(t)}
\ \text{and}\ \pi_{j^\star(t)} > p_{t-1},\\[4pt]
p_{t-1}, & \text{otherwise,}
\end{cases}
\qquad
r^{\mathrm{prog}}_t = \indicator{p_t > p_{t-1}} .
\label{eq:reward}
\end{equation}

The agent receives reward only when it reaches a higher confident progress value. Low-confidence matches and repeated visits to the same progress level receive zero reward. Rewarding progress change rather than raw similarity provides a stable dense signal and reduces reward hacking from static high-similarity states or noisy score fluctuations. We use RARM as a drop-in reward with DrQ-v2~\citep{yarats2021mastering} for training from scratch in simulation, and with DSRL~\citep{wagenmaker2025steering} for steering initial noise for a VLA policy in the real world.

%===============================================================================

\section{Experiments}
\label{sec:experiments}

We evaluate RARM in both simulation and real-world settings. Our experiments are designed to answer four questions: \begin{enumerate} \item Can RARM provide dense rewards for RL training from scratch? \item How does RARM compare with representative reference-based and VLM-based progress reward models? \item Can RARM support policy fine-tuning in noisy real-world settings? 
\item Which components are important for reliable progress rewards? \end{enumerate}

\subsection{Comparisons and ablations}
\label{sec:sim_comparisons}

\begin{figure*}[t]
  \centering
  \setlength{\fboxsep}{0pt}%
  {\normalsize\bfseries Short horizon tasks}\\[3pt]
  \makebox[\textwidth]{%
    \panel{0.228\textwidth}{MT: coffee pull}{\includegraphics[width=0.228\textwidth,height=0.16\textwidth]{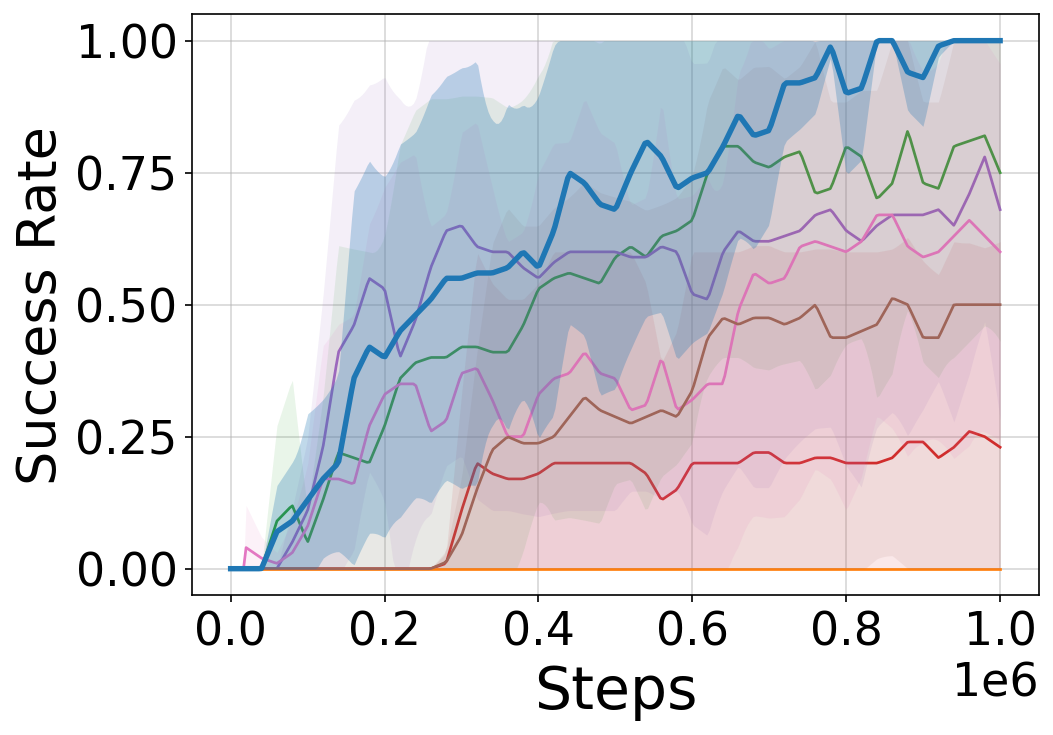}}
    \panel{0.198\textwidth}{MT: button press wall}{\includegraphics[width=0.198\textwidth,height=0.16\textwidth]{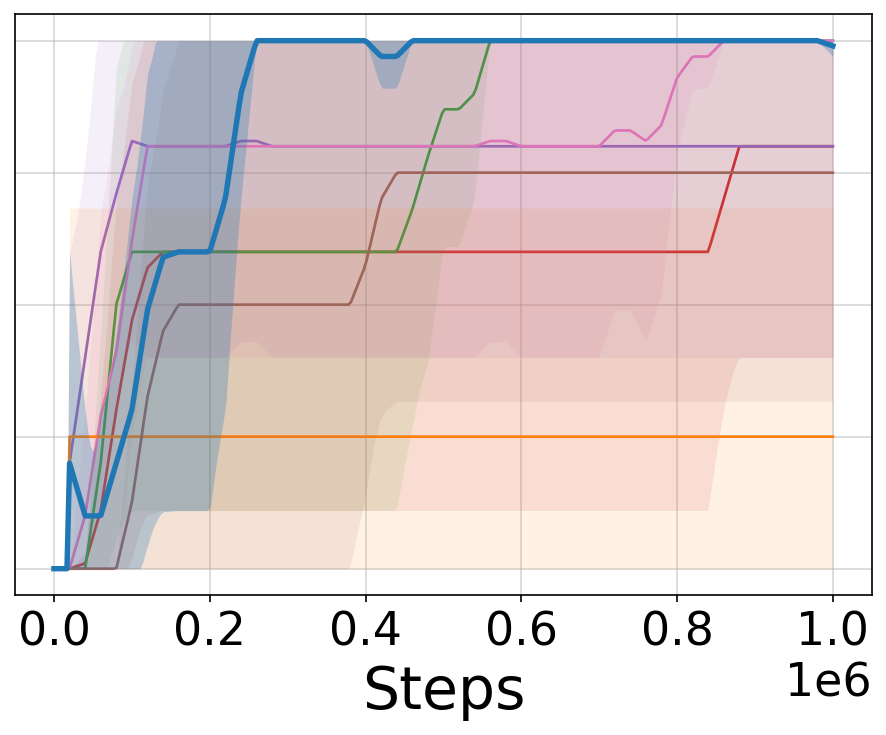}}
    \panel{0.198\textwidth}{MT: drawer open}{\includegraphics[width=0.198\textwidth,height=0.16\textwidth]{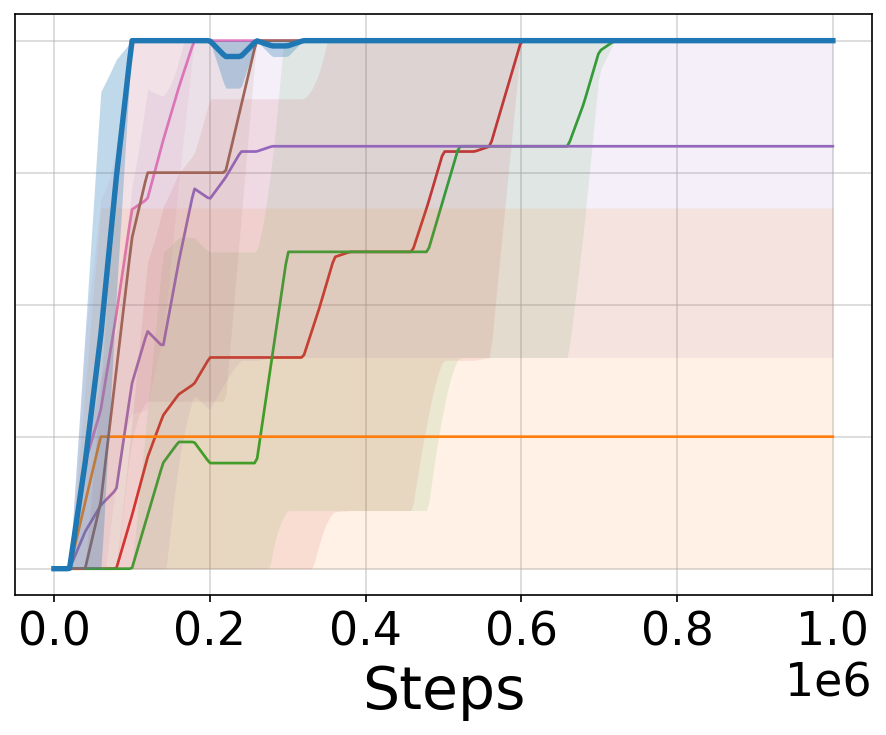}}
    \panel{0.198\textwidth}{MT: soccer}{\includegraphics[width=0.198\textwidth,height=0.16\textwidth]{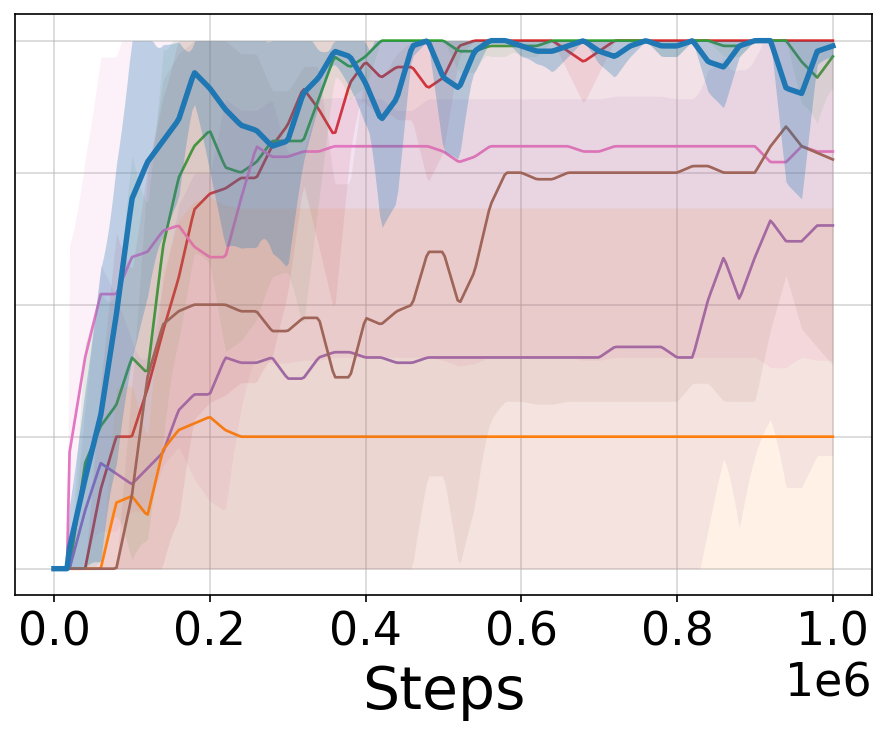}}
    \panel{0.198\textwidth}{LIB10: book in caddy}{\includegraphics[width=0.198\textwidth,height=0.16\textwidth]{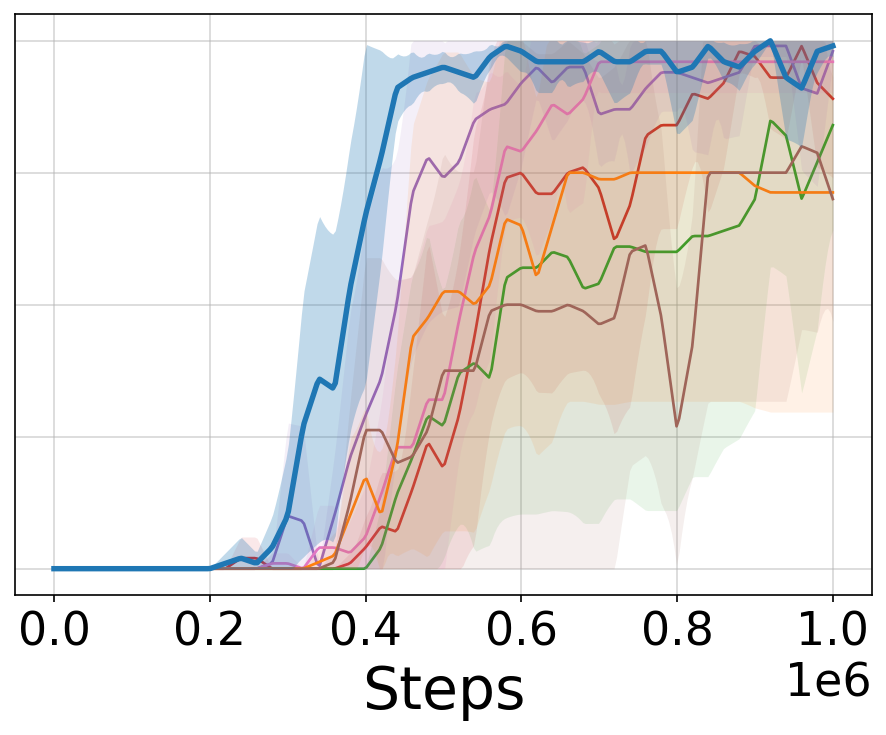}}
  }\\[5pt]
  {\normalsize\bfseries Long horizon tasks}\\[3pt]
  \makebox[\textwidth]{%
    \panel{0.278\textwidth}{LIB10: stove + pot}{\includegraphics[width=0.278\textwidth,height=0.16\textwidth]{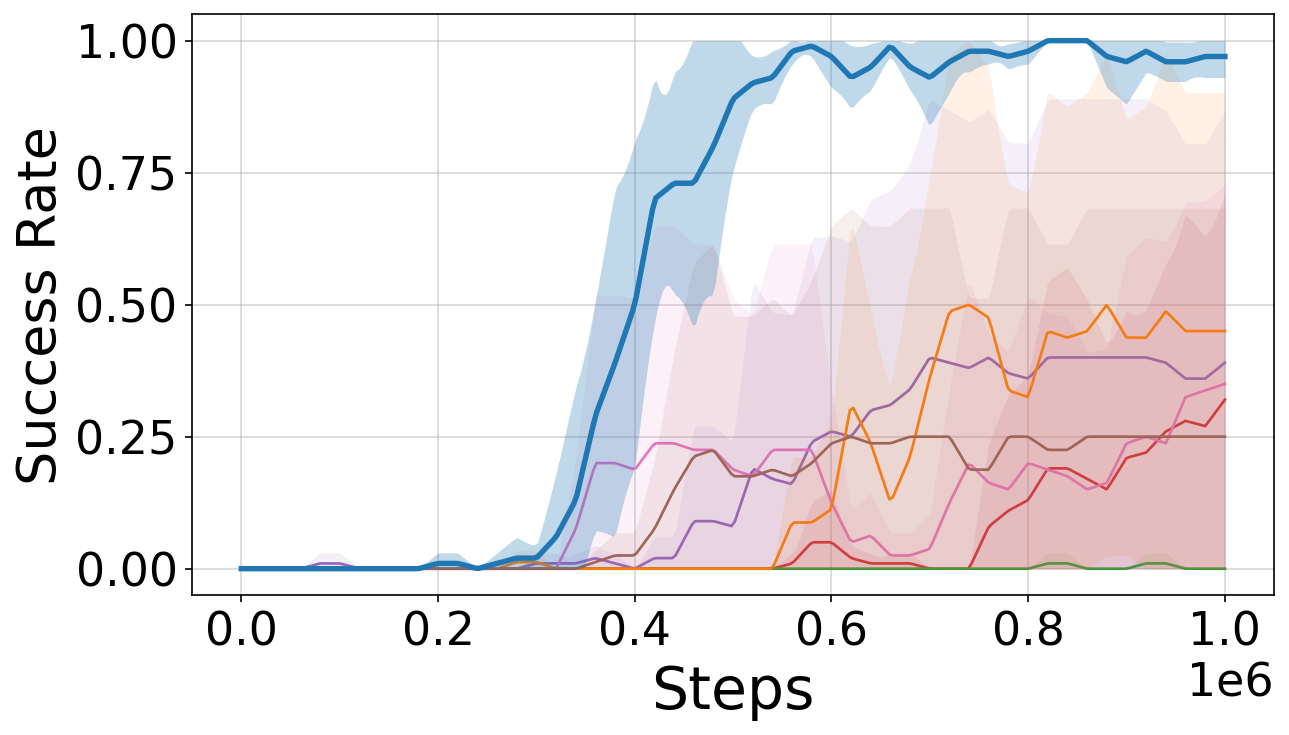}}
    \panel{0.248\textwidth}{LIB10: bowl in drawer + close}{\includegraphics[width=0.248\textwidth,height=0.16\textwidth]{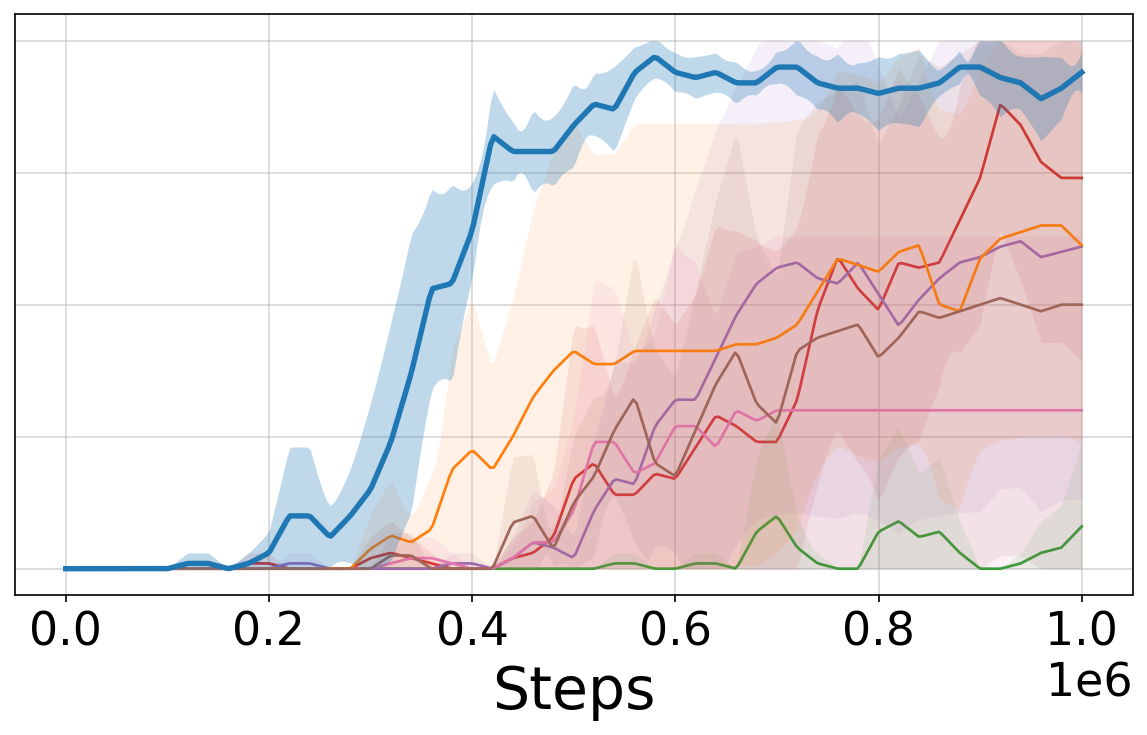}}
    \panel{0.248\textwidth}{LIB10: mug + pudding}{\includegraphics[width=0.248\textwidth,height=0.16\textwidth]{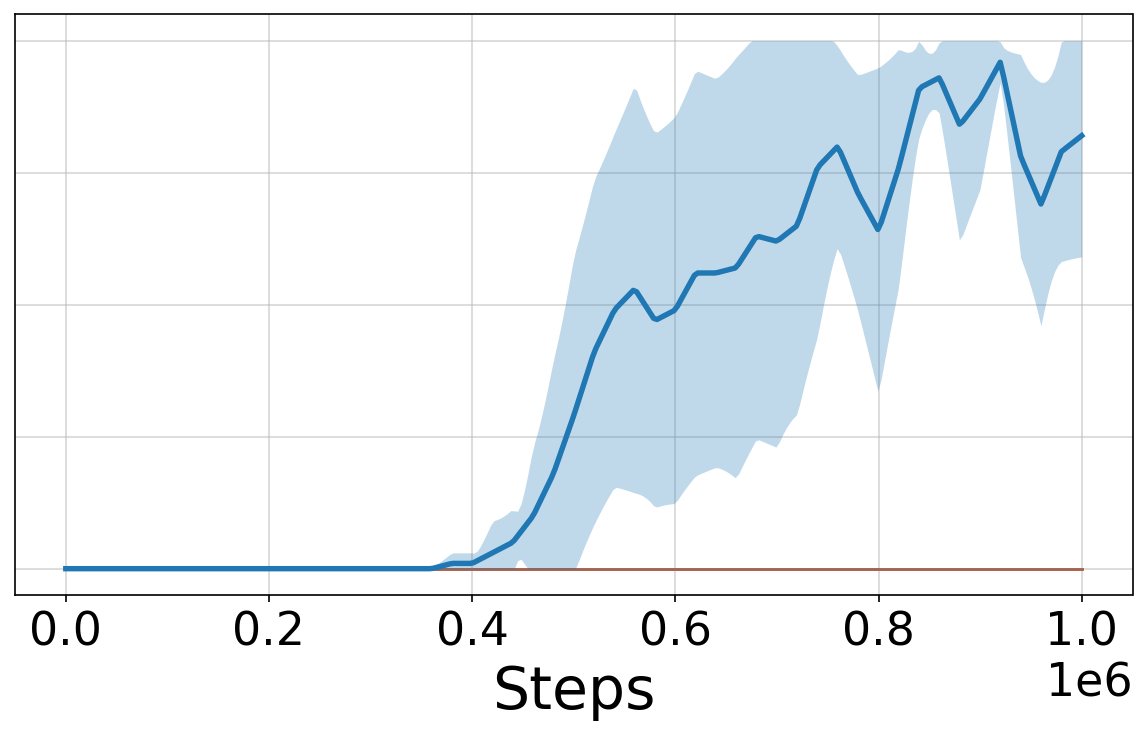}}
    \panel{0.248\textwidth}{LIB10: mug in microwave + close}{\includegraphics[width=0.248\textwidth,height=0.16\textwidth]{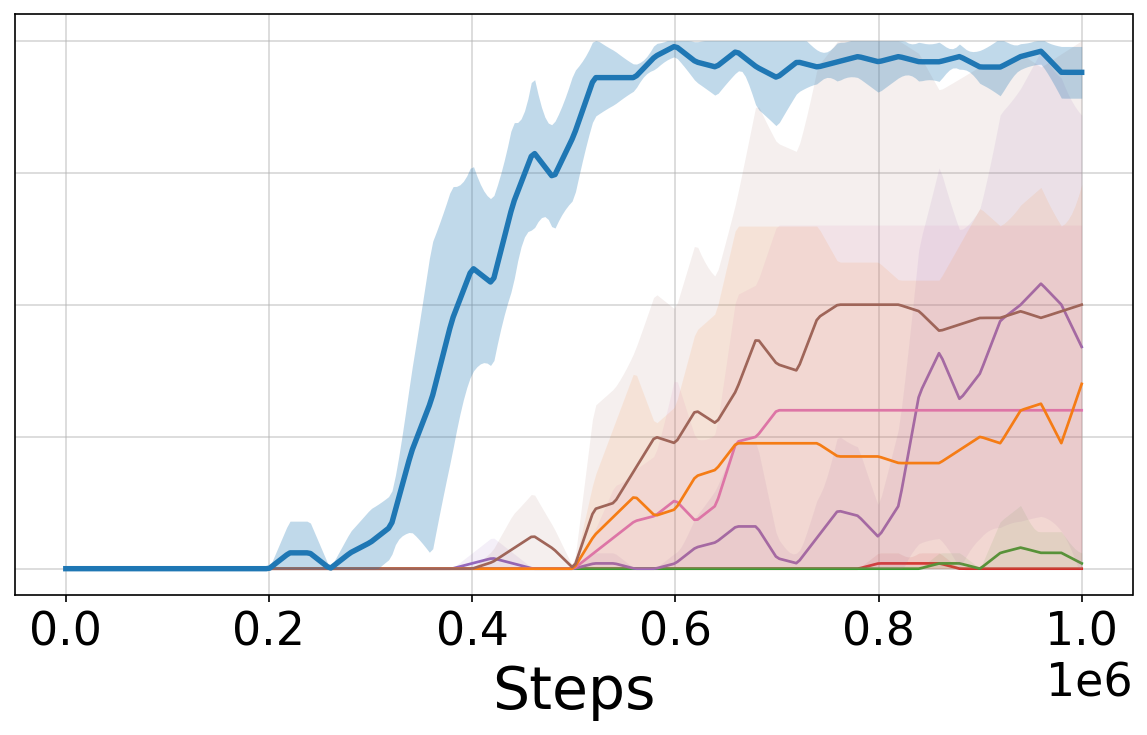}}
  }\\[5pt]
  \makebox[\textwidth][c]{\scriptsize%
    \leg{cRARM}{RARM (Ours)}\hspace{1em}%
    \leg{cRoboCLIP}{RoboCLIP}\hspace{1em}%
    \leg{cTemporalOT}{TemporalOT}\hspace{1em}%
    \leg{cGVL}{GVL}\hspace{1em}%
    \leg{cRoboMeter}{RoboMeter}\hspace{1em}%
    \leg{cRoboDopamine}{RoboDopamine}\hspace{1em}%
    \leg{cAblationSim}{AblationSim}%
  }
  \caption{Success rate over 1M environment steps across 4 MetaWorld and 5 LIBERO-10 tasks (same 5 seeds). Baselines consists of: VLM-based (GVL, RoboMeter, RoboDopamine), refernced-based (RoboCLIP, TemporalOT), and an ablation (AblationSim). Our method, RARM (blue), achieves consistently better performance across all tasks. }
  \vspace{-0.6cm}
  \label{fig:learning_curve}
\end{figure*}

We evaluate RL training from scratch on four MetaWorld tasks and five LIBERO-10 tasks, covering both short- and long-horizon manipulation. All methods are trained with DrQ-v2~\citep{yarats2021mastering} for $10^6$ environment steps using five random seeds. We evaluate every $10$k steps with $10$ rollouts per seed. Each method uses its own predicted dense reward together with the task success signal. Experiments are run on NVIDIA RTX 5090 and GH200 GPUs. 

\textbf{Baselines.} We compare RARM against representative reference-based and VLM-based reward models. The reference-based baselines are \textbf{RoboCLIP}~\citep{sontakke2023roboclip}, which scores a rollout by video-language embedding similarity to a reference, and \textbf{TemporalOT}~\citep{fu2024robot}, which aligns a robot trajectory to an expert demonstration using temporally constrained optimal transport. The VLM-based baselines are \textbf{GVL}~\citep{ma2024vision}, \textbf{RoboDopamine}~\citep{tan2025robo}, and \textbf{RoboMeter}~\citep{liang2026robometer}, which predict task progress from visual observations using large vision-language models or VLM-trained reward models. We also include \textbf{Similarity Reward}, an ablation of RARM that removes progress-change computation and directly uses the maximum rollout--reference similarity as the step reward. 

\begin{wrapfigure}{r}{0.4\linewidth} \centering \vspace{-1em} \panel{0.4\textwidth}{LIB10: stove + pot}{ \includegraphics[ width=0.35\textwidth, height=0.2\textwidth ]{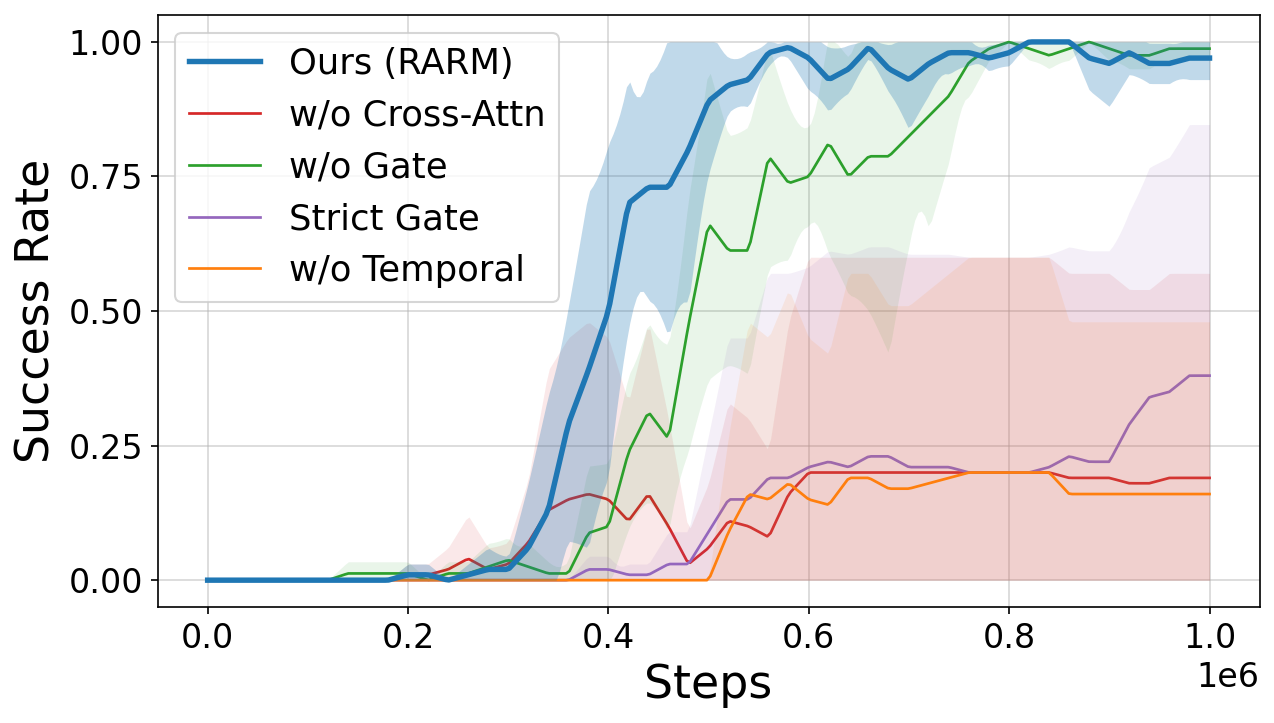} } \panel{0.4\textwidth}{MT: drawer open}{ \includegraphics[ width=0.35\textwidth, height=0.2\textwidth ]{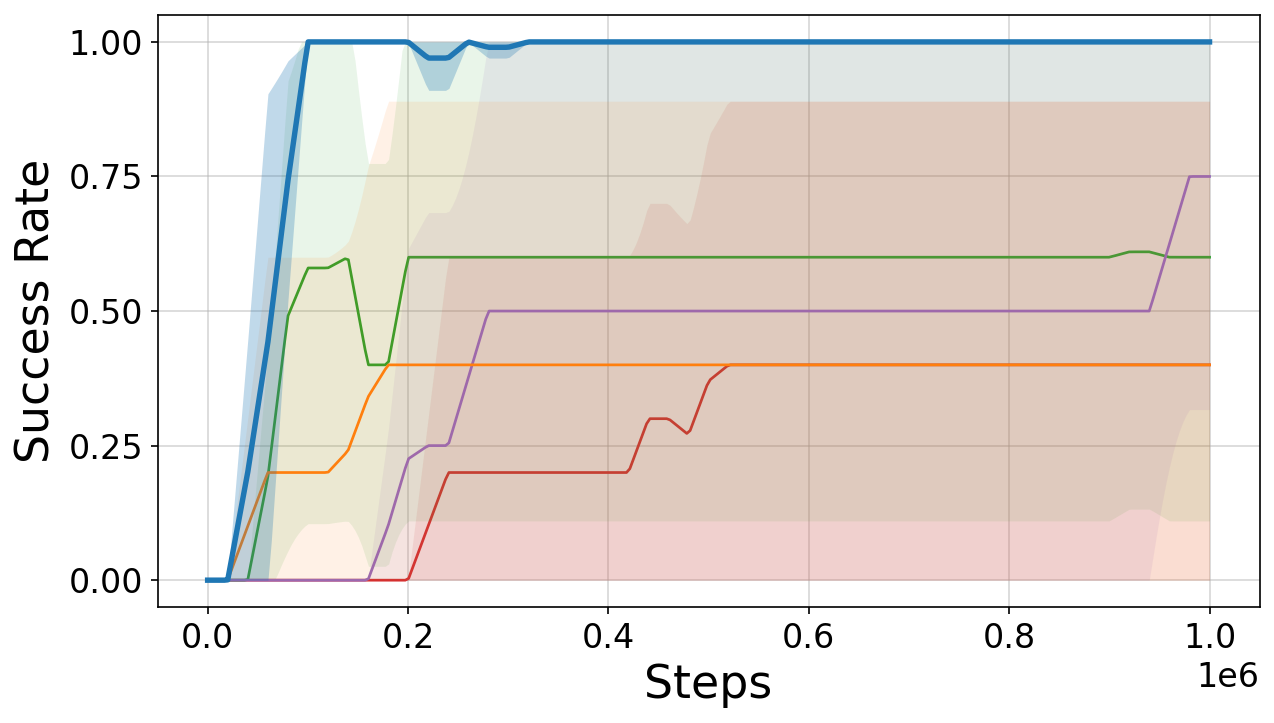} } \caption{\textbf{Ablation.} Success rate over 1M steps for four component ablations on one LIBERO and one MetaWorld task, averaged over the same five seeds.} \label{fig:ablation} \vspace{-2em} \end{wrapfigure}

\textbf{Main results.} Figure~\ref{fig:learning_curve} summarizes the simulation learning curves and addresses Q1--Q2: RARM provides dense rewards that support RL training from scratch and outperform representative reference-based and VLM-based progress reward models, it achieves the strongest overall performance across both benchmark suites. On short-horizon MetaWorld tasks, several baselines can eventually learn useful policies, but RARM reaches high success faster and more consistently. On long-horizon LIBERO-10 tasks, reference-similarity baselines such as RoboCLIP and TemporalOT struggle, suggesting that direct trajectory or embedding alignment is brittle when rollouts differ in speed, include pauses, or require intermediate corrections. VLM-based baselines provide dense semantic progress estimates, but their predictions are often noisy or over-optimistic on partially completed rollouts; we analyze this failure mode qualitatively in Sec.~\ref{sec:Qualitative Comparisons}. In contrast, RARM uses reference matching only to localize progress and rewards confidence-gated progress changes, which yields stable dense rewards even for long-horizon tasks.

\textbf{Ablation studies.} The full comparison includes \textbf{Similarity Reward} on all nine tasks, testing whether raw rollout--reference similarity alone is sufficient. Its weaker performance shows the importance of converting confident matches into progress-change rewards. We further ablate the confidence gate, cross-attention comparator, and temporal processor on one long-horizon LIBERO task and one short-horizon MetaWorld task (Fig.~\ref{fig:ablation}). Removing the gate admits unreliable matches, whereas an overly strict gate suppresses useful progress updates. Removing cross-attention weakens rollout--reference matching, while removing the temporal processor discards motion-order information needed to distinguish forward from reversed motion. All variants perform worse than the full model, showing that confidence gating, learned comparison, and temporal reasoning jointly support reliable progress estimation.

% The full comparison includes \textbf{Similarity Reward} on all nine tasks, directly testing whether raw rollout--reference similarity is sufficient. Its weaker performance shows that RARM's gains do not come merely from comparing a rollout to a reference; the similarity score must be converted into a confidence-gated progress update. To further examine the core components, we ablate the confidence threshold and the cross-attention comparator on one representative long-horizon LIBERO task and one short-horizon MetaWorld task (Fig.~\ref{fig:ablation}). A low threshold accepts unreliable matches and can over-reward false progress, while a high threshold rejects useful matches and makes the reward sparse. Removing cross-attention weakens rollout--reference comparison and degrades progress localization. These results answer Q4 and show that both confidence gating and learned clip comparison are important for reliable progress rewards.

\subsection{Qualitative Comparisons}
\label{sec:Qualitative Comparisons}

\begin{figure}[t]
    \centering
    \includegraphics[width=\linewidth,trim={0 0 0 0},clip]{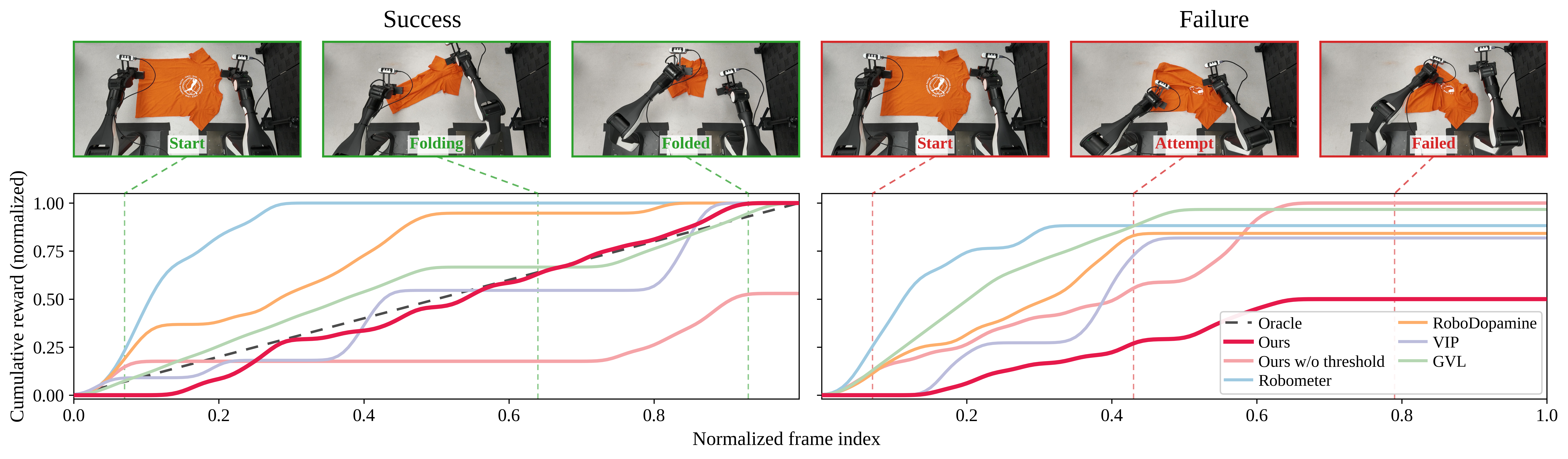}
    \caption{\textbf{Cumulative reward on successful and failed cloth-folding rollouts.} For each reward model, both curves are normalized by the larger of its final cumulative rewards on the successful and failed rollouts. \emph{For this visualization, all models use the same rule: $+1$ only when estimated progress exceeds its previous maximum, and $0$ otherwise. The curves are therefore nondecreasing by construction; they show accumulated rewards, not raw progress predictions or each model's native reward formulation.} RARM additionally gates progress updates by confidence. Its failure-to-success reward ratio is $0.50$, compared with GVL ($0.97$), RoboMeter ($0.88$), RoboDopamine ($0.84$), and VIP ($0.82$). The VLM-based baselines also accumulate reward early in the successful rollout. Removing RARM's confidence gate increases the ratio to approximately $1.89$.}
    \label{fig:qualitative_comparisons}
    \vspace{-1em}
\end{figure}

We further analyze Q4 using paired successful and failed cloth-folding rollouts (Fig.~\ref{fig:qualitative_comparisons}). To compare progress estimates under a consistent rule, each model receives a unit reward only when its estimate exceeds its previous maximum; we plot the cumulative sum. RARM applies its confidence gate before accepting an update. RARM accumulates reward throughout the successful rollout and saturates earlier on failure, whereas the baselines assign comparatively high cumulative reward to failure or accumulate reward prematurely. Removing the gate increases the failure-to-success reward ratio to approximately $1.89$, supporting the role of confidence filtering in this example.

\subsection{Real-world experiments}

\noindent\textbf{Setup.} We evaluate RARM as a drop-in dense reward for DSRL~\citep{wagenmaker2025steering} on top of the $\pi_0$ VLA~\citep{black2024pi_0}, deployed on an AIRBOT Play dual-arm platform. Four real-world tasks span the practical difficulty spectrum: two single-arm short-horizon tasks (\emph{Pick Eraser}, \emph{Drawer Open}) and two dual-arm long-horizon tasks (\emph{Bimanual Hand-over} and the three-stage \emph{Bimanual Clothes Folding}); full task setups, success criteria, and filmstrip visualisations are given in Appendix. For each task we compare four settings --- the base $\pi_0$, DSRL driven by a sparse binary success reward (\emph{BinR}), DSRL driven by the RoboMeter reward model~\citep{liang2026robometer} (\emph{RoboM}), and DSRL driven by our RARM reward (\emph{Ours}) --- at training budgets of $\{30, 60, 90\}$ on-robot rollouts, and evaluate every resulting checkpoint over $30$ rollouts (Table~\ref{tab:rollout_success}).

\noindent\textbf{Results.} Table~\ref{tab:rollout_success} answers Q3: RARM can serve as a dense reward for real-world DSRL training under noisy, low-data conditions. The base $\pi_0$ clears at most a third of rollouts on the easiest task and fewer than one in ten on both long-horizon tasks, confirming that off-the-shelf VLA imitation alone is insufficient on this hardware. Across the two short-horizon tasks and the bimanual hand-over, RARM dominates both baselines at every training budget, improving on the sparse-binary reward by $3$--$10$ rollouts out of $30$ --- and already at the smallest budget ($19/30$ vs.\ $16/30$ on Drawer Open, $17/30$ vs.\ $9/30$ on Pick Eraser, and $14/30$ vs.\ $5/30$ on Hand-over) --- so the dense progress signal lets DSRL extract a useful gradient from substantially fewer real-robot episodes. RoboMeter improves over BinR in some short-horizon settings, but falls below it on Hand-over at every budget ($5/30$ versus $14/30$ at $90$ training episodes), showing
that dense rewards can hinder learning when progress estimates are unreliable. The gap widens decisively on Clothes Folding, the hardest task in the suite: BinR barely improves on the base policy ($7/30$ at $90$ rollouts versus $2/30$ without DSRL), because a clean three-stage fold almost never occurs during exploration and so the binary signal leaves the policy with virtually no useful gradient. RARM instead recovers a dense progress signal even on partial folds (Stage~1 alone, Stage~1 plus the $45^\circ$ rotation, etc.) and reaches $25/30$ at the same budget --- more than $3\times$ BinR and $2.8\times$ RoboMeter. At the smallest budget no method has yet moved this task appreciably ($4/30$ for both RARM and RoboMeter), so the advantage emerges only once enough partial-fold experience has accumulated. This is the regime where reference-anchored reward shaping helps most: when binary success is vanishing during exploration, a continuous progress signal turns an otherwise untrainable task into a tractable one.

\begin{wraptable}{l}{0.75\textwidth}   % l = left side; text wraps on the right
    \centering
    \vspace{-0.5em}
    \setlength{\tabcolsep}{3pt}
    \renewcommand{\arraystretch}{1.1}
    \caption{Real-world success rates over 30 evaluation rollouts before and after 30/60/90 DSRL training episodes. \textbf{w/o DSRL:} base VLA policy. \textbf{BinR:} DSRL with sparse binary success reward. \textbf{RoboM:} DSRL with the RoboMeter reward. \textbf{Ours:} DSRL with the proposed RARM reward.
    }
    \label{tab:rollout_success}
    \resizebox{0.7\textwidth}{!}{%
    \begin{tabular}{l c ccc ccc ccc}
    \toprule
    \multirow{2}{*}{Task} & \multirow{2}{*}{w/o DSRL}
      & \multicolumn{3}{c}{30 rollouts} & \multicolumn{3}{c}{60 rollouts} & \multicolumn{3}{c}{90 rollouts} \\
    \cmidrule(lr){3-5}\cmidrule(lr){6-8}\cmidrule(lr){9-11}
      & & BinR& RoboM& Ours& BinR& RoboM& Ours& BinR& RoboM& Ours\\
    \midrule
    Drawer open     & 10/30& 16/30& 17/30& \textbf{19/30}& 19/30& 18/30& \textbf{23/30}& 23/30& 20/30& \textbf{29/30} \\
    Pick eraser     & 5/30& 9/30& 13/30& \textbf{17/30}& 15/30& 17/30& \textbf{21/30}& 20/30& 20/30& \textbf{25/30} \\
    Hand over       & 2/30& 5/30& 3/30& \textbf{14/30}& 12/30& 4/30& \textbf{22/30}& 14/30& 5/30& \textbf{23/30} \\
    Clothes folding & 2/30& 2/30& \textbf{4/30}& \textbf{4/30}& 3/30& 9/30& \textbf{12/30}& 7/30& 9/30& \textbf{25/30} \\
    \bottomrule
    \vspace{-0.5em}
    \end{tabular}}
\end{wraptable}

\noindent\textbf{Robustness to lighting and environmental noise.}
Real-world rollouts include randomized object poses, illumination
changes, and background clutter. The same RARM checkpoint supports
policy learning across all four tasks without task-specific
reward-model fine-tuning. Its training on generic videos with color
jitter and random cropping is intended to improve tolerance to
appearance changes (Section~\ref{sec:approach:contrastive}).
Controlled experiments further examine glare, color shifts,
occlusion, and viewpoint changes
(Appendix~\ref{sec: RARM Robustness}). We also evaluate synthetic
reference demonstrations as alternative progress anchors
(Appendix~\ref{append:video_gen}).

\begin{wraptable}{l}{0.5\textwidth}
    \centering
    \caption{\textbf{LIBERO-10 recovery.}
    Success rates (\%) across four seeds under
    mid-episode arm displacements.}
    \label{tab:error_recovery}
    \setlength{\tabcolsep}{3pt}

    \resizebox{\linewidth}{!}{%
    \begin{tabular}{@{}lcccc@{}}
        \toprule
        \textbf{Model}
        & \textbf{None}
        & \textbf{5\,cm}
        & \textbf{10\,cm}
        & \textbf{20\,cm} \\
        \midrule
        GVL
        & $45.1\pm8.8$
        & $38.7\pm8.4$
        & $26.3\pm6.0$
        & $17.5\pm4.5$ \\
        RoboDopamine
        & $36.7\pm8.1$
        & $31.9\pm7.4$
        & $23.2\pm5.4$
        & $15.1\pm3.4$ \\
        RoboMeter
        & $48.8\pm6.2$
        & $40.5\pm4.9$
        & $34.7\pm4.0$
        & $23.7\pm3.7$ \\
        TemporalOT
        & $22.6\pm5.6$
        & $18.6\pm4.9$
        & $15.8\pm4.1$
        & $14.1\pm4.0$ \\
        \textbf{RARM}
        & $\boldsymbol{95.6\pm1.0}$
        & $\boldsymbol{90.9\pm1.5}$
        & $\boldsymbol{78.9\pm2.2}$
        & $\boldsymbol{48.2\pm4.8}$ \\
        \bottomrule
    \end{tabular}%
    }
\end{wraptable}

\subsection{Multiple Solution Modes and Scope}
\label{sec:multi-modal}

RARM uses a single successful demonstration to bootstrap RL policy learning, without assuming that one demonstration covers every valid solution mode. The reference anchors progress estimation, while all-to-all clip matching allows variation in execution speed and recovery from deviations. We evaluate recovery through mid-episode interventions: displacing the robot arm in a random direction and resetting object positions to require re-grasping. Table~\ref{tab:error_recovery} reports the arm-displacement results; object-reset examples are provided on the project webpage. RARM achieves the highest success rate under every intervention magnitude, supporting recovery beyond the demonstrated trajectory. These results demonstrate flexibility around the reference execution, while coverage of distinct solution strategies remains an open direction.

\section{Conclusion}
\label{sec:conclusion}

We presented \emph{Reference-Anchored Reward Model} (RARM), a lightweight reward model that derives confidence-gated progress rewards from a single successful demonstration. Rather than using raw similarity as reward, RARM localizes rollout progress along the reference and rewards only confident forward progress, reducing false-positive reward assignment during RL. Trained with self-supervised contrastive learning, RARM requires no task-specific reward supervision, subtask annotations, failure data, or large-scale robot datasets.

Experiments in simulation and the real world show that RARM provides fast, low-overhead reward queries and outperforms VLM-based, trajectory-alignment, and raw-similarity reward baselines. Ablations confirm the importance of confidence gating and learned clip comparison. While RARM assumes access to a representative reference demonstration, its effectiveness suggests that reference-anchored, confidence-gated progress rewards offer a practical route toward reliable RL post-training for robotic manipulation.

\section{Limitations}
RARM provides a lightweight and data-efficient way to derive progress rewards from one reference demonstration, but it also has several limitations. First, because RARM relies on visual clip comparison rather than explicit semantic reasoning, its robustness may degrade under large embodiment, viewpoint, or scene changes. 

Second, like other vision-based reward models, RARM may struggle with fine-grained manipulation details that are difficult to infer from RGB observations alone. Small contact changes, grasp stability, force interaction, or subtle object deformation can be critical for success but visually ambiguous. Incorporating additional modalities such as depth, tactile sensing may help improve.

Third, RARM provides progress based on reference trajetory, which makes repetitive or cyclic tasks challenging, since visually similar states may appear multiple times at different progress levels. Extending RARM to handle repeated stages, multi-reference demonstrations, or explicit memory over longer horizons is an important direction for future work.

%===============================================================================

% \clearpage
% The acknowledgments are automatically included only in the final and preprint versions of the paper.
% \acknowledgments{This work was supported by MOE AcRF Tier 1 funding (24-1234-P0001), the NUS Presidential Young Professorship, and the Google Gift Fund.}

\acknowledgments{
This work was supported by MOE AcRF Tier 1 funding
(24-1234-P0001), the NUS Presidential Young Professorship,
the Google Gift Fund, and the NVIDIA Academic Grant Program.
Additional support was provided through Swiss AI Initiative
project ID 19, with a small grant from the ETH Domain and
computational resources on Alps at the Swiss National
Supercomputing Centre (CSCS).
}

%===============================================================================
\clearpage
\bibliography{example}  % .bib
\clearpage
\newpage
\appendix
\section{Simulation Environments}
\begin{longtable}{m{0.2\textwidth} m{0.10\textwidth} m{0.28\textwidth} m{0.28\textwidth}}
\caption{Simulation tasks used in our evaluation. The image column is reserved for task visualizations.} \\
\toprule
\textbf{Simulation Environment} & \textbf{Task} & \textbf{Image} & \textbf{Description} \\
\midrule
\endfirsthead
\toprule
\textbf{Simulation Environment} & \textbf{Task} & \textbf{Image} & \textbf{Description} \\
\midrule
\endhead
\midrule
\multicolumn{4}{r}{\textit{Continued on next page}} \\
\endfoot
\bottomrule
\endlastfoot
MetaWorld (MT50) & Task 7
& \includegraphics[width=3.2cm,height=2.4cm]{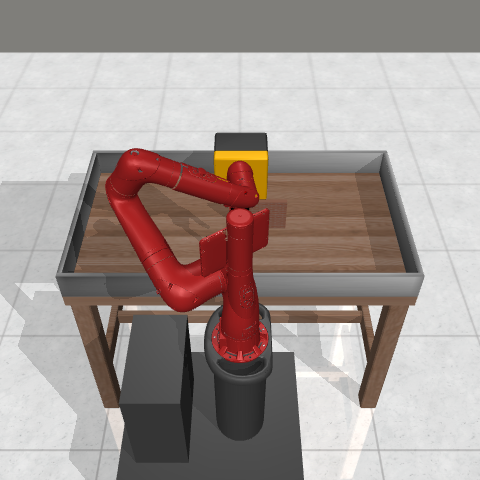}
& Bypass a wall and press a button. \\

MetaWorld (MT50) & Task 9
& \includegraphics[width=3.2cm,height=2.4cm]{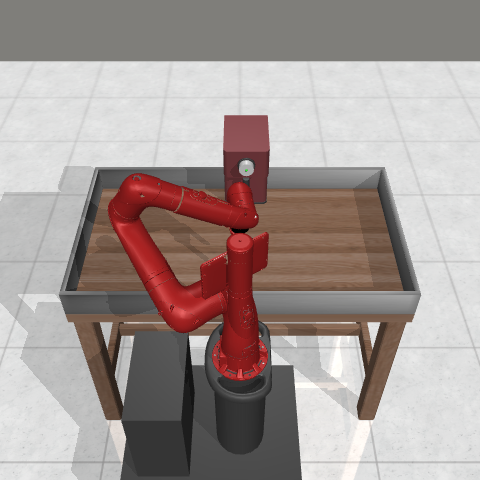}
& Pull a mug from a coffee machine. \\

MetaWorld (MT50) & Task 18
& \includegraphics[width=3.2cm,height=2.4cm]{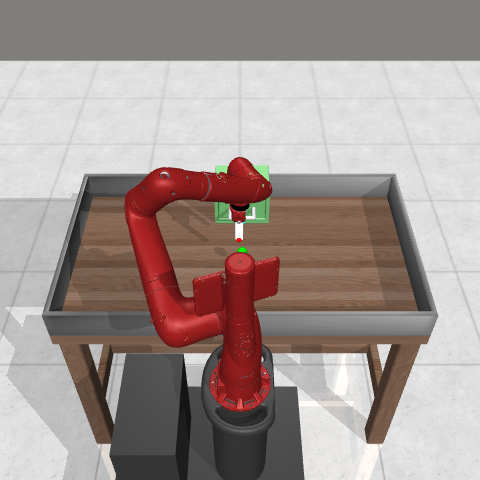}
& Open a drawer. \\

MetaWorld (MT50) & Task 43
& \includegraphics[width=3.2cm,height=2.4cm]{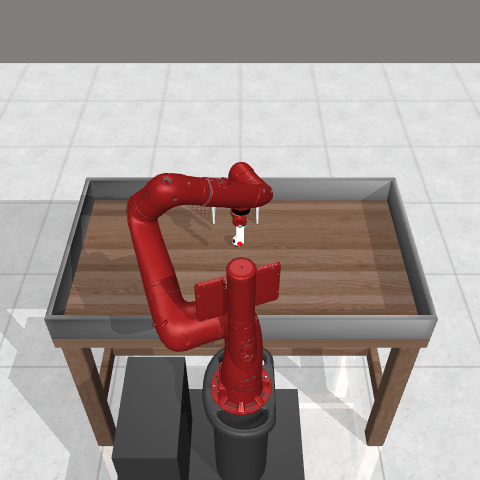}
& Kick a soccer into the goal. \\

% MetaWorld (MT50) & Task 49
% & \includegraphics[width=3.4cm,height=2.4cm]{figure/metaworld_task49_window_close.png}
% & Push and close a window. \\
\midrule
\midrule
LIBERO-10 (libero-long) & Task 2
& \includegraphics[width=3.2cm,height=2.4cm]{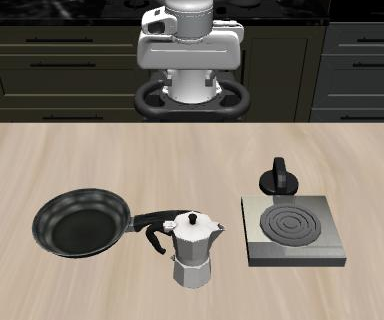}
& Turn on the stove and put the moka pot on it. \\

LIBERO-10 (libero-long) & Task 3
& \includegraphics[width=3.2cm,height=2.4cm]{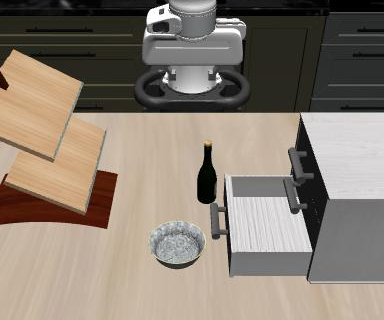}
& Put the black bowl in the bottom drawer of the cabinet and close it. \\

LIBERO-10 (libero-long) & Task 5
& \includegraphics[width=3.2cm,height=2.4cm]{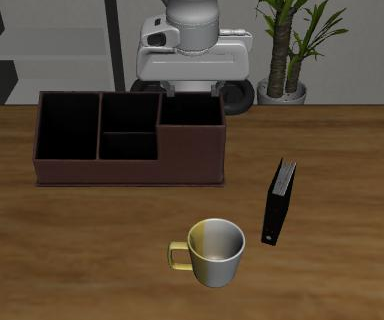}
& Pick up the book and place it in the back compartment of the caddy. \\

LIBERO-10 (libero-long) & Task 6
& \includegraphics[width=3.2cm,height=2.4cm]{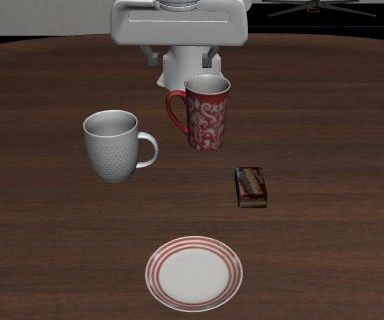}
& Put the white mug on the plate and put the chocolate pudding to the right of the plate. \\

LIBERO-10 (libero-long) & Task 9
& \includegraphics[width=3.2cm,height=2.4cm]{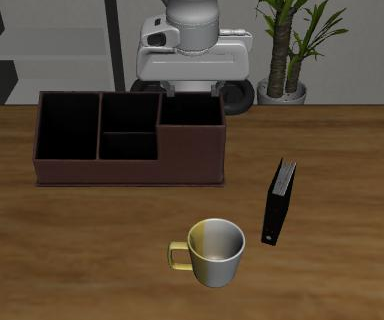}
& Put the yellow and white mug in the microwave and close it. \\

\end{longtable}

% \clearpage
\section{Real-world Tasks}
\label{sec:real-world-tasks}

\subsection*{Hardware and Training Setup}

All four real-world tasks are executed on a single \textbf{AIRBOT Play} dual-arm platform. Each arm has 6 revolute joints plus a 1-DoF parallel-jaw gripper (maximum opening width $0.07$\,m), giving a 7-dimensional per-arm joint + gripper action space. The two single-arm tasks (\emph{Pick Eraser}, \emph{Drawer Open}) use only the left arm of the same robot; the two dual-arm tasks (\emph{Hand-over}, \emph{Clothes Folding}) use both arms in coordination. Observations come from three Intel RealSense cameras streaming $640{\times}480$ color at $30$\,fps --- one top-mounted scene camera and two wrist-mounted cameras (one per arm) --- downsampled to $128{\times}128$ as input to the $\pi_0$ policy and to $384{\times}384$ as input to the RARM comparator.

The base $\pi_0$ policy predicts an action chunk of $50$ future steps and re-infers every $25$ executed steps; actions are streamed to the robot at \textbf{$20$\,Hz}. RARM rewards are scored at clip boundaries with a stride of $4$ environment steps ($\approx 5$\,Hz reward updates), which keeps GPU usage low without sacrificing temporal resolution of the progress signal.

\noindent\textbf{A note on figure orientation.} For visualisation we record the filmstrips in this section from a fixed high-angle camera placed in front of the robot (i.e.\ facing the robot from the opposite side of the workspace). Because of this viewpoint the on-screen left/right is mirrored relative to the robot's own frame: \emph{the gripper that appears on the right side of each frame is the robot's left arm, and the gripper on the left side is the robot's right arm.} Whenever the text below refers to ``left arm'' or ``right arm'' we mean the robot's own arms, not the on-screen positions.

\subsection{Task 1: Pick Eraser from Box (Single-Arm, Short-Horizon)}

\begin{figure}[h]
  \centering
  \makebox[\linewidth]{%
    \includegraphics[width=0.19\linewidth]{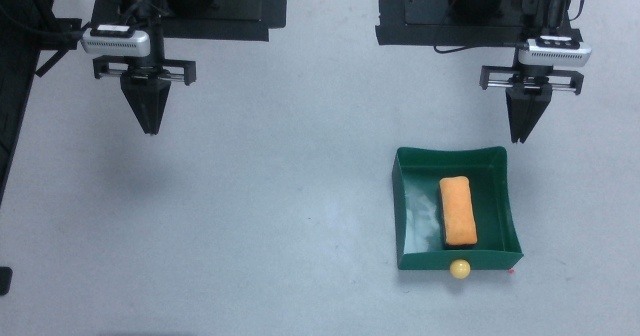}\hfill
    \includegraphics[width=0.19\linewidth]{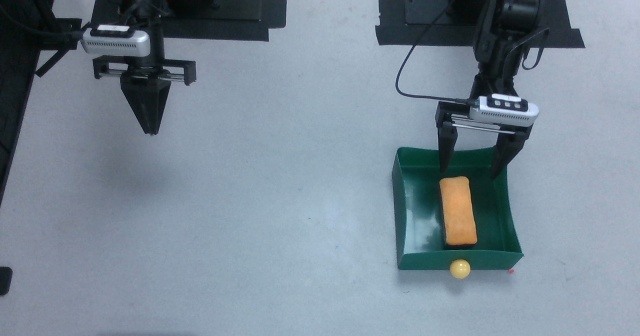}\hfill
    \includegraphics[width=0.19\linewidth]{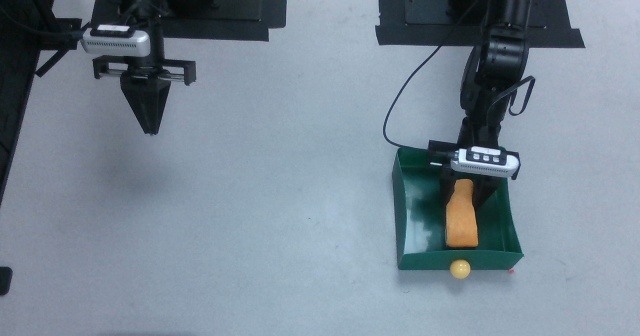}\hfill
    \includegraphics[width=0.19\linewidth]{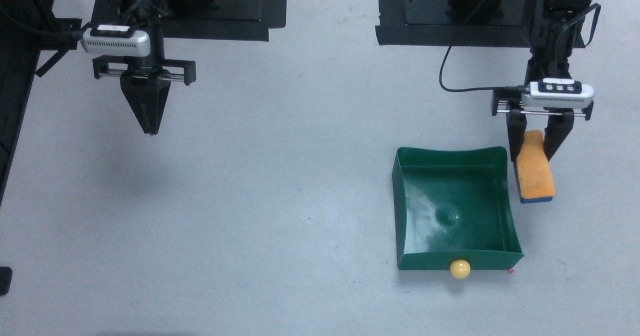}\hfill
    \includegraphics[width=0.19\linewidth]{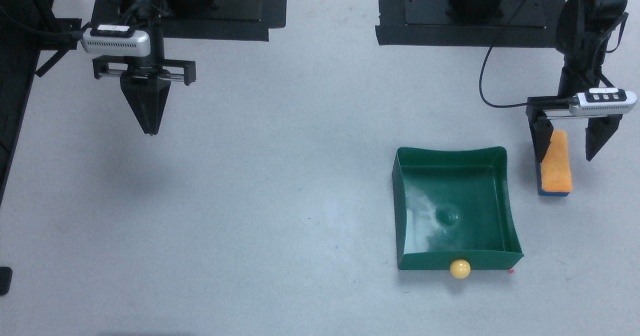}%
  }\\[2pt]
  \makebox[\linewidth]{%
    \parbox[t]{0.19\linewidth}{\centering\footnotesize Approach\\ box}\hfill
    \parbox[t]{0.19\linewidth}{\centering\footnotesize Descend\\ over eraser}\hfill
    \parbox[t]{0.19\linewidth}{\centering\footnotesize Close\\ gripper}\hfill
    \parbox[t]{0.19\linewidth}{\centering\footnotesize Lift clear\\ of rim}\hfill
    \parbox[t]{0.19\linewidth}{\centering\footnotesize Place on\\ tabletop}%
  }
  \caption{\textbf{Task 1 --- Pick Eraser from Box.} Filmstrip of one successful rollout sampled at uniform intervals; action-phrase labels appear under each frame.}
  \label{fig:real-task1-eraser}
\end{figure}

\noindent\textbf{Task description.} A single-arm short-horizon pick-and-place task. The robot must reach into a green storage box, grasp an orange-and-yellow blackboard eraser whose pose inside the box is randomised across episodes, lift it cleanly past the box rim, and release it onto the open tabletop. The task tests precise vision-based grasping inside a contained workspace and a clean release-then-retract motion at the end.

\noindent\textbf{Setup \& metrics.} A rollout counts as a success if, at the timeout, the eraser lies on the tabletop fully outside the box footprint and the gripper has released and retracted. Evaluated over $30$ rollouts (Table~\ref{tab:rollout_success}, row ``Pick eraser'').

\subsection{Task 2: Open Red Drawer (Single-Arm, Short-Horizon)}

\begin{figure}[h]
  \centering
  \makebox[\linewidth]{%
    \includegraphics[width=0.19\linewidth]{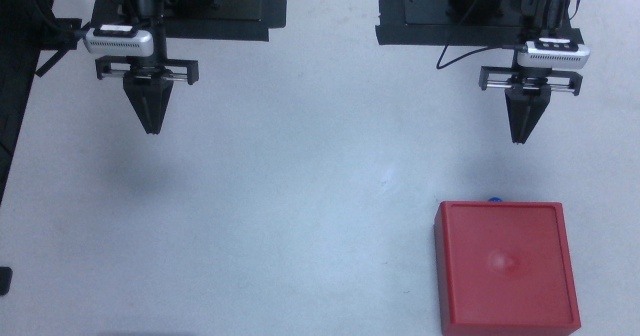}\hfill
    \includegraphics[width=0.19\linewidth]{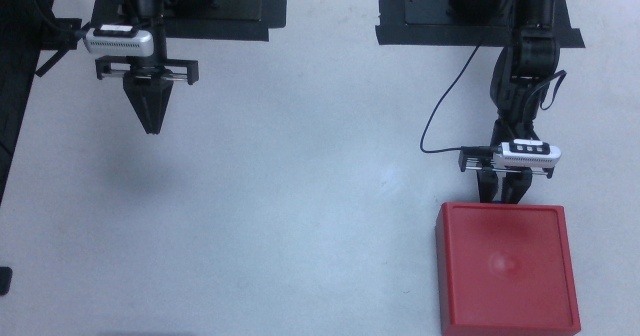}\hfill
    \includegraphics[width=0.19\linewidth]{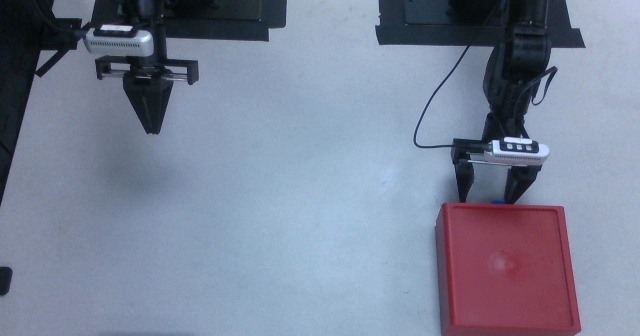}\hfill
    \includegraphics[width=0.19\linewidth]{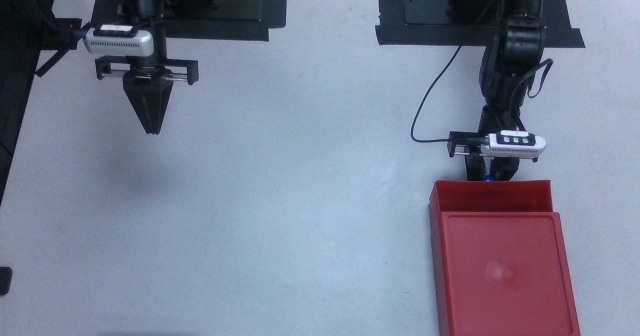}\hfill
    \includegraphics[width=0.19\linewidth]{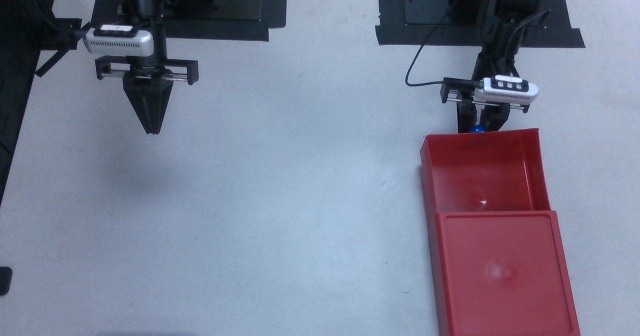}%
  }\\[2pt]
  \makebox[\linewidth]{%
    \parbox[t]{0.19\linewidth}{\centering\footnotesize Approach\\ drawer}\hfill
    \parbox[t]{0.19\linewidth}{\centering\footnotesize Align with\\ handle}\hfill
    \parbox[t]{0.19\linewidth}{\centering\footnotesize Close on\\ blue handle}\hfill
    \parbox[t]{0.19\linewidth}{\centering\footnotesize Pull\\ halfway}\hfill
    \parbox[t]{0.19\linewidth}{\centering\footnotesize Pull fully\\ open}%
  }
  \caption{\textbf{Task 2 --- Open Red Drawer.} Filmstrip of one successful rollout sampled at uniform intervals; action-phrase labels appear under each frame.}
  \label{fig:real-task2-drawer}
\end{figure}

\noindent\textbf{Task description.} A single-arm short-horizon articulated-object manipulation task. The robot must localise the blue plastic handle on the red drawer (mounted on the drawer face that points toward the robot, hence not visible in the top-down filmstrip view above), close the gripper firmly on it, and execute a straight pulling motion along the drawer rail until the drawer is fully extended. The task probes handle localisation accuracy and the policy's ability to maintain a stable grasp throughout a constrained linear motion against the rail's reaction forces.

\noindent\textbf{Setup \& metrics.} A rollout counts as a success if, at the timeout, the drawer is extended by more than ${\sim}70\%$ of its travel range and the gripper is still attached to the handle. Evaluated over $30$ rollouts (Table~\ref{tab:rollout_success}, row ``Drawer open'').

\subsection{Task 3: Bimanual Hand-over (Dual-Arm, Long-Horizon)}

\begin{figure}[h]
  \centering
  \makebox[\linewidth]{%
    \includegraphics[width=0.19\linewidth]{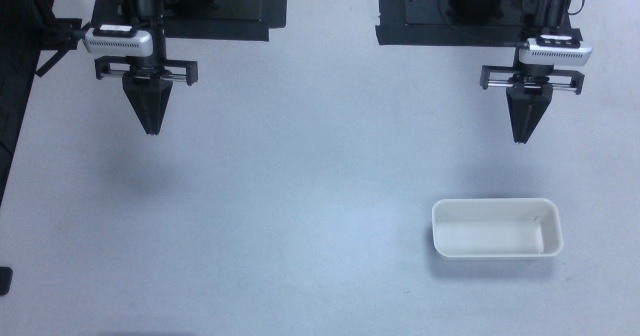}\hfill
    \includegraphics[width=0.19\linewidth]{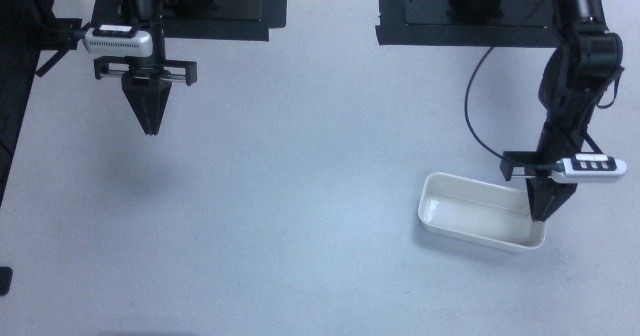}\hfill
    \includegraphics[width=0.19\linewidth]{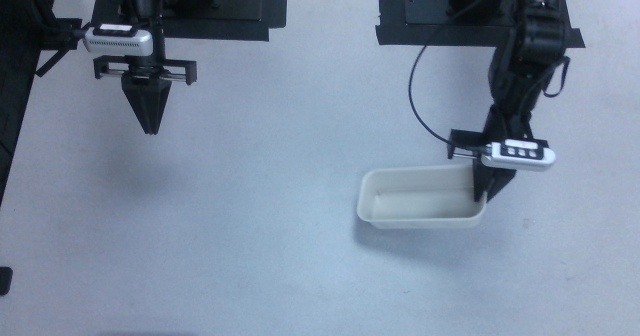}\hfill
    \includegraphics[width=0.19\linewidth]{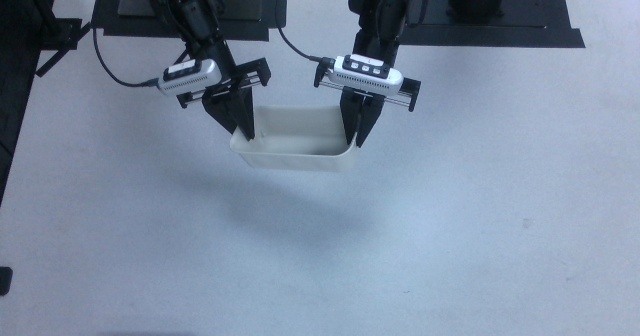}\hfill
    \includegraphics[width=0.19\linewidth]{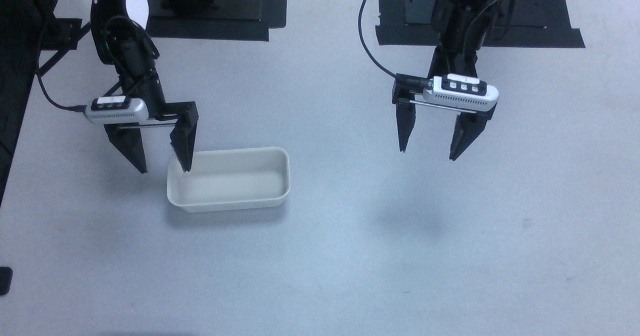}%
  }\\[2pt]
  \makebox[\linewidth]{%
    \parbox[t]{0.19\linewidth}{\centering\footnotesize Both arms\\ at home}\hfill
    \parbox[t]{0.19\linewidth}{\centering\footnotesize Left grasps\\ container}\hfill
    \parbox[t]{0.19\linewidth}{\centering\footnotesize Lift to\\ hand-over pose}\hfill
    \parbox[t]{0.19\linewidth}{\centering\footnotesize Right closes,\\ left releases}\hfill
    \parbox[t]{0.19\linewidth}{\centering\footnotesize Right places\\ on table}%
  }
  \caption{\textbf{Task 3 --- Bimanual Hand-over.} Filmstrip of one successful rollout sampled at uniform intervals; action-phrase labels appear under each frame.}
  \label{fig:real-task3-handover}
\end{figure}

\noindent\textbf{Task description.} A dual-arm long-horizon coordination task. The left arm picks the off-white container off the tabletop, raises it to a mid-air hand-over pose, and must hold steady while the right arm reaches in and closes around the same object; only after the right gripper is closed does the left arm release. The right arm then transports the container down to the tabletop and releases. The task stresses cross-arm temporal synchronisation: an early release by the left arm drops the container, a late close by the right arm collides with the container, and the reward signal must credit hand-over progress as a coordinated event rather than rewarding either arm in isolation.

\noindent\textbf{Setup \& metrics.} The episode timeout is chosen to comfortably cover the longest rollout, including the mid-air dwell during the handshake. A rollout counts as a success if, by the timeout, the container has been placed on the tabletop at the right arm's release pose and both grippers are clear of the object. Evaluated over $30$ rollouts (Table~\ref{tab:rollout_success}, row ``Hand over'').

\subsection{Task 4: Bimanual Clothes Folding (Dual-Arm, Long-Horizon, Multi-Stage)}

\begin{figure}[h]
  \centering
  \makebox[\linewidth]{%
    \includegraphics[width=0.19\linewidth]{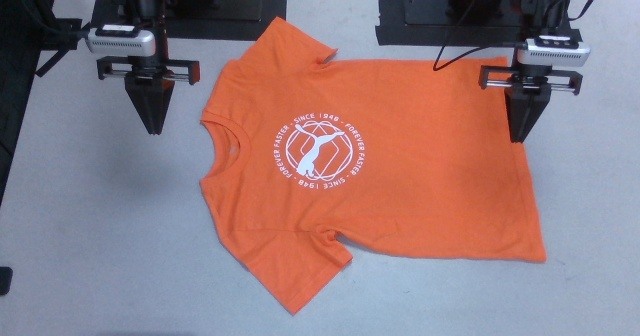}\hfill
    \includegraphics[width=0.19\linewidth]{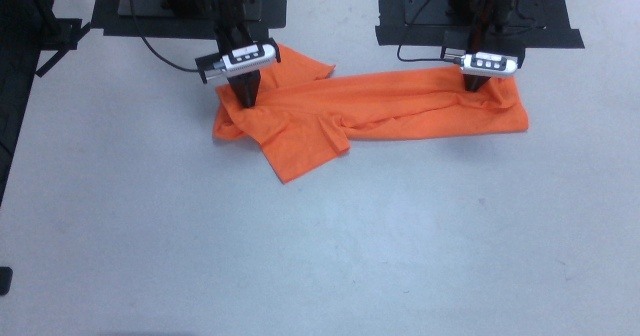}\hfill
    \includegraphics[width=0.19\linewidth]{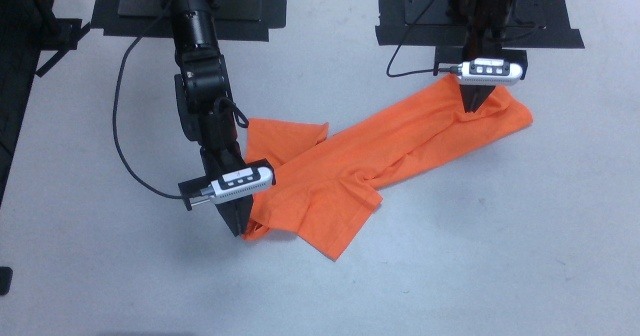}\hfill
    \includegraphics[width=0.19\linewidth]{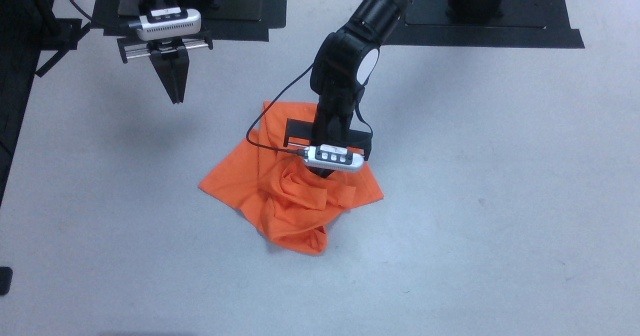}\hfill
    \includegraphics[width=0.19\linewidth]{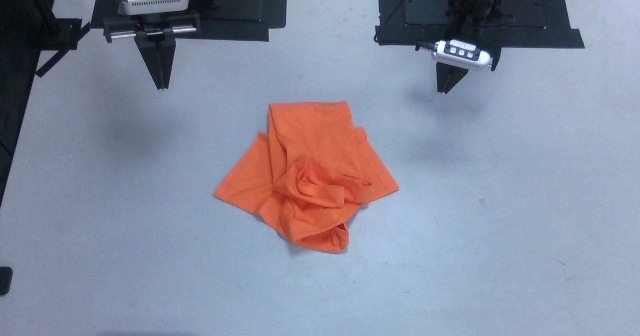}%
  }\\[2pt]
  \makebox[\linewidth]{%
    \parbox[t]{0.19\linewidth}{\centering\footnotesize Flat orange\\ shirt}\hfill
    \parbox[t]{0.19\linewidth}{\centering\footnotesize Bimanual\\ first fold}\hfill
    \parbox[t]{0.19\linewidth}{\centering\footnotesize Cooperative\\ $45^{\circ}$ rotation}\hfill
    \parbox[t]{0.19\linewidth}{\centering\footnotesize Left pinches\\ near side}\hfill
    \parbox[t]{0.19\linewidth}{\centering\footnotesize Final folded\\ shape}%
  }
  \caption{\textbf{Task 4 --- Bimanual Clothes Folding.} Filmstrip of one successful rollout sampled at uniform intervals; action-phrase labels appear under each frame.}
  \label{fig:real-task4-fold}
\end{figure}

\noindent\textbf{Task description.} A dual-arm long-horizon multi-stage cloth-manipulation task on a flat orange shirt. \emph{Stage~1}: both arms simultaneously pinch one side of the shirt and fold it across to the opposite side. \emph{Stage~2}: the two arms cooperatively rotate the partially folded shirt by $45^{\circ}$ on the tabletop to re-align it. \emph{Stage~3}: the left arm alone pinches the side now closest to it and folds it across to the corresponding opposite side, yielding the final folded shape. This is the most demanding evaluation task in our suite: it combines bi-manual coordination, deformable-object dynamics, and three causally-ordered stages, which jointly make it an effective stress test of the reward model's progress signal under long-horizon multi-stage behaviour.

\noindent\textbf{Setup \& metrics.} A rollout counts as a success only if all three stages complete in order and the final folded shape lies within the target footprint on the tabletop; partial folds (e.g.\ Stage~1 only) are recorded as failures. Evaluated over $30$ rollouts (Table~\ref{tab:rollout_success}, row ``Clothes folding'').

% \clearpage
\section{Heuristic on Confidence Gating}

The comparator is trained contrastively, so its scores are calibrated only in a
relative sense: what counts as a strong match depends on which clip is being
matched. A clip covering a distinctive motion attracts higher scores from everything than one covering a slow or visually uneventful stretch of the demonstration. A single global threshold would therefore be too permissive for the former and too strict for the latter.

The reference demonstration already carries the information needed to set this scale. Comparing it against itself once, offline, reveals how strong a match
each of its own clips is able to attract:
\begin{equation}
m_j \;=\; \max_{k} \, g_\theta(r_j, r_k).
\label{eq:selfsim}
\end{equation}
We read $m_j$ as the score clip $j$ receives when it is matched as well as
anything in this demonstration can match it. A rollout clip is accepted as a
confident match to $r_j$ when it falls within a fixed slack of that value:
\begin{equation}
\delta_j \;=\; m_j - \Delta .
\label{eq:gate}
\end{equation}

The slack $\Delta$ is a single scalar shared by every reference clip
($\Delta = 0.5$); all per-clip variation in the gate comes from $m_j$, which is
measured rather than chosen. The gate therefore asks whether a rollout clip
matches $r_j$ about as well as the reference matches itself there, rather than
whether it exceeds any particular similarity value -- which is what allows one
rule to apply across tasks without recalibration.

$\Delta$ is the only free quantity in the gate, and Fig.~\ref{fig:ablation}
ablates it at both extremes. Large $\Delta$ admits every nearest-clip match and
reduces the estimate to ungated arg-max localisation (\emph{w/o Gate}); small
$\Delta$ admits only near-duplicates, so progress rarely updates and the reward
approaches the sparse setting (\emph{Strict Gate}).

\section{Settings}
\subsection{Other Hyperparameters}
Remaining hyperparameters follow the public DrQ-v2~\citep{yarats2021mastering}
and DSRL~\citep{wagenmaker2025steering} implementations. Within each benchmark
they are identical across reward models, so performance differences reflect the
reward signal rather than the RL backbone.

% Appendix: submission-time hyperparameters.
% Requires: \usepackage{booktabs, multirow, array}
% Drop the tables you want into the appendix; cut the rest.

% =====================================================================
% Table X: DrQ-v2 optimization hyperparameters
% =====================================================================
\begin{table}[h]
\centering
\caption{DrQ-v2 optimization hyperparameters. Identical across MetaWorld and LIBERO.}
\label{tab:hp-drqv2-opt}
\begin{tabular}{ll}
\toprule
\textbf{Parameter} & \textbf{Value} \\
\midrule
Optimizer                          & Adam (separate for encoder, actor, critic) \\
Learning rate                      & $1 \times 10^{-4}$ \\
Batch size                         & 256 \\
Discount factor $\gamma$           & 0.99 \\
$n$-step return                    & 3 \\
Critic target EMA $\tau$           & 0.01 \\
Update frequency                   & every 2 env steps \\
Seed frames (no updates before)    & 4000 \\
Exploration steps (uniform random) & 2000 \\
Exploration noise schedule         & $\sigma(t) = \mathrm{linear}(1.0, 0.1, 500{,}000)$ \\
Exploration noise clip             & 0.3 \\
\bottomrule
\end{tabular}
\end{table}

\begin{table}[h]
\centering
\caption{DrQ-v2 network architecture. Conv2d layers are specified by output channels, kernel size, and stride. The actor and critic have separate projection trunks; the two Q-functions share the critic trunk.}
\label{tab:drqv2_architecture}
\resizebox{\linewidth}{!}{%
\begin{tabular}{lll}
\toprule
\textbf{Component} & \textbf{Layer / Operation} & \textbf{Output Shape} \\
\midrule
Data augmentation & Replicate padding $+4$, random crop & $(N_{\mathrm{ch}},84,84)$ \\
Shared CNN encoder & Conv2d$(32,3\!\times\!3,2)$ + ReLU & $(32,41,41)$ \\
 & Conv2d$(32,3\!\times\!3,1)$ + ReLU & $(32,39,39)$ \\
 & Conv2d$(32,3\!\times\!3,1)$ + ReLU & $(32,37,37)$ \\
 & Conv2d$(32,3\!\times\!3,1)$ + ReLU & $(32,35,35)$ \\
 & Flatten & $39{,}200$ \\
\midrule
Actor trunk & Linear$(39{,}200\!\rightarrow\!50)$ + LayerNorm + Tanh & $50$ \\
Critic trunk & Linear$(39{,}200\!\rightarrow\!50)$ + LayerNorm + Tanh & $50$ \\
\midrule
Actor head & Linear$(50\!\rightarrow\!1024)$ + ReLU & $1024$ \\
 & Linear$(1024\!\rightarrow\!1024)$ + ReLU & $1024$ \\
 & Linear$(1024\!\rightarrow\!\mathcal A_{\mathrm{dim}})$ + Tanh & $\mathcal A_{\mathrm{dim}}$ \\
 & Truncated normal with scheduled standard deviation & $\mathcal A_{\mathrm{dim}}$ \\
\midrule
Each critic head ($\times 2$) & Linear$(50+\mathcal A_{\mathrm{dim}}\!\rightarrow\!1024)$ + ReLU & $1024$ \\
 & Linear$(1024\!\rightarrow\!1024)$ + ReLU & $1024$ \\
 & Linear$(1024\!\rightarrow\!1)$ & $1$ \\
\bottomrule
\end{tabular}%
}
\end{table}

% =====================================================================
% Table X: Observation and replay buffer
% =====================================================================
% \begin{table}[t]
% \centering
% \caption{Observation stack and replay buffer settings.}
% \label{tab:hp-obs-replay}
% \begin{tabular}{lll}
% \toprule
% \textbf{Parameter} & \textbf{MetaWorld} & \textbf{LIBERO} \\
% \midrule
% Frame stack                  & 3                  & 3 \\
% Channels per frame           & 3 (agentview RGB)  & 6 (agentview + wrist RGB) \\
% Stacked observation shape    & $9 \times 84 \times 84$ & $18 \times 84 \times 84$ \\
% \bottomrule
% \end{tabular}
% \end{table}

\begin{center}
\captionof{table}{Observation stack and replay buffer settings.}
\label{tab:hp-obs-replay}
\begin{tabular}{@{}lll@{}}
\toprule
\textbf{Parameter} & \textbf{MetaWorld} & \textbf{LIBERO} \\
\midrule
Frame stack               & 3 & 3 \\
Channels per frame        & 3 (agentview RGB) & 6 (agentview + wrist RGB) \\
Stacked observation shape & $9 \times 84 \times 84$ & $18 \times 84 \times 84$ \\
\bottomrule
\end{tabular}
\end{center}

% \subsection{Detailed Network Architecture}

% \begin{figure}[h]
% 	\centering
% 	% \includegraphics[width=\linewidth]{figures/architecture.pdf}
% 	\framebox[\linewidth]{\parbox[c][3.0cm][c]{0.92\linewidth}{\centering\textbf{[Figure~2 placeholder --- model architecture]}\\[2pt] frames $\rightarrow$ DINOv3 ViT-B/16 + LoRA $\rightarrow$ temporal processor (spatial pooling + temporal RoPE + cross-frame attention) $\rightarrow$ cross-attention comparator $\rightarrow$ comparison head $\rightarrow$ scalar $g_\theta$}}
% 	\caption{\textbf{Comparator architecture.} A frozen DINOv3 encoder with LoRA adapters embeds each frame; a temporal processor injects time information and mixes frames; an asymmetric cross-attention module compares the query clip against a candidate clip; and a small head emits a single scalar similarity.}
% 	\label{fig:arch}
% \end{figure}

\subsection{Data used for RARM Training}
\label{sec:data used for RARM training}

RARM is trained entirely with a self-supervised contrastive objective
(Sec.~\ref{sec:approach:contrastive}) and requires no task labels, reward annotations, or robot-specific demonstrations. The training corpus is drawn from DAVIS~\cite{Caelles_arXiv_2019} and MOSE~\cite{MOSE, ding2025mosev2}, video object segmentation datasets covering a wide variety of real-world scenes: people, animals, vehicles, and outdoor activities. We filter out near-static sequences by computing the per-frame IoU of the annotated object mask between consecutive frames and discarding any sequence whose mean inter-frame IoU exceeds a fixed threshold. The remaining sequences provide diverse natural motion and appearance variation, encouraging the comparator to learn temporal dynamics that generalise beyond any specific robot or laboratory setting.

% \clearpage

\section{Additional Results}
\label{Additional Results}

\subsection{video generation for RM generalization}
\label{append:video_gen}

\begin{wrapfigure}{r}{0.4\linewidth}
    \vspace{-10pt}
    \centering
    \begin{minipage}{\linewidth}
      (a)\\[-2pt]
      \includegraphics[width=\linewidth]{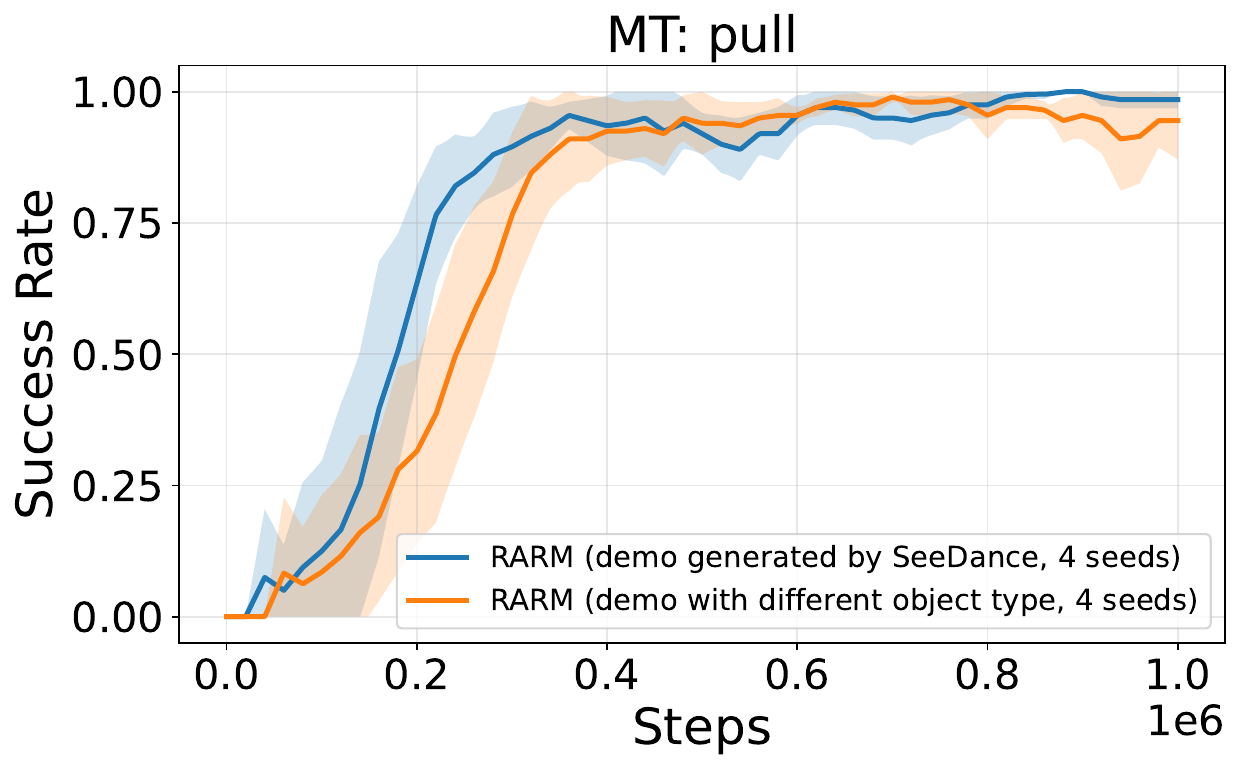}

      \vspace{4pt}

      (b)\\[-2pt]
      \begin{minipage}{0.38\linewidth}
        \centering
        \includegraphics[width=\linewidth]{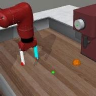}
      \end{minipage}
      \hfill
      \begin{minipage}{0.48\linewidth}
        \centering
        \includegraphics[width=\linewidth]{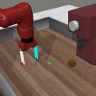}
      \end{minipage}
    \end{minipage}
    \caption{Generated reference demonstration on the MetaWorld pull task.
      (a)~Learning curves comparing RARM using a real reference with a different object instance, and synthetic references
      generated by SeeDance~2.0.
      (b)~Example frames from two synthetic trajectories generated by
      SeeDance~2.0 from an initial scene image and task instruction.}
    \label{fig:seedance_demo}
    \vspace{-25pt}
\end{wrapfigure}

RARM requires a single reference demonstration for each new task configuration. While this requirement is substantially lighter than collecting large task-specific demonstration datasets, the need for even a single human-collected reference can still limit scalability when tasks scale combinatorially or when human data collection is expensive. We therefore study whether RARM can successfully leverage non-expert, visually mismatched, or synthetically generated reference videos as approximate progress anchors.

We evaluate this capability on the MetaWorld pull task across two distinct challenge settings:

1. \textbf{Robustness to Novel Objects}: The reference video depicts the correct underlying task structure but features an object instance with a completely different appearance (color, shape, and texture) from the one encountered by the robot during RL training. This setting evaluates whether RARM relies on pixel-level comparison or can instead abstract task-level motion correspondences across distinct objects. The resulting learning curve of 2 different object types with 2 seeds for each is illustrated in orange in Figure~\ref{fig:seedance_demo} (a).

2. \textbf{Learning from Synthetic References}: Given only a static initial scene image and a textual language instruction describing the target task, we utilize SeeDance 2.0 to synthesize a successful execution video. This generated video is utilized as the sole reference trajectory for RARM, completely eliminating human capture from the loop. The corresponding learning curve is shown in blue in Figure~\ref{fig:seedance_demo} (a).

As shown in Figure~\ref{fig:seedance_demo}, policy training remains highly effective when relying on both visually mismatched and synthetically generated references, achieving asymptotic success rates and sample efficiency comparable to a human-collected baseline.  These results support the interpretation that RARM does not require pixel-perfect alignment with the reference video (more robustness analysis can be found in \ref{sec: RARM Robustness}). Instead, the cross-attention comparator treats the reference trajectory as a coarse temporal scaffold for progress localization, while the calibrated confidence gate naturally suppresses the visual artifacts, hallucinations, and structural inconsistencies inherent to generative video models. This validation points to a highly scalable paradigm for visual reward design: downstream policies can be optimized using automated pipelines where text-to-video foundation models generate approximate progress anchors on demand.

\graphicspath{{figure/}}

\newlength{\hgap}\setlength{\hgap}{6pt}
\newlength{\vgap}\setlength{\vgap}{6pt}

\newlength{\fivew}
\setlength{\fivew}{\dimexpr(\textwidth - 4\hgap - 2pt)/5\relax}

\newlength{\gh}\setlength{\gh}{3.3cm}
\newlength{\titleh}\setlength{\titleh}{0.7cm}
\newlength{\frameh}\setlength{\frameh}{\dimexpr\gh - \titleh - \vgap\relax}

\newlength{\leftw}\setlength{\leftw}{\dimexpr3\fivew + 2\hgap\relax}
\newlength{\leftcx}\setlength{\leftcx}{\dimexpr\leftw/2\relax}
\newlength{\stripw}\setlength{\stripw}{\dimexpr5\fivew + 4\hgap\relax}
\newlength{\largeregion}\setlength{\largeregion}{\dimexpr2\fivew + \hgap\relax}

\newlength{\fy}\setlength{\fy}{\dimexpr\titleh + \vgap\relax}
\newlength{\colbx}\setlength{\colbx}{\dimexpr\fivew + \hgap\relax}
\newlength{\colcx}\setlength{\colcx}{\dimexpr2\fivew + 2\hgap\relax}
\newlength{\graphx}\setlength{\graphx}{\dimexpr3\fivew + 3\hgap\relax}
\newlength{\graphcx}\setlength{\graphcx}{\dimexpr\graphx + \fivew + \hgap/2\relax}

\newcommand{\imgbox}[1]{%
  \begin{tikzpicture}
    \node[inner sep=0pt]{\includegraphics[width=\fivew,keepaspectratio]{figure/#1}};
  \end{tikzpicture}%
}

\newcommand{\bigrow}[2]{%
  \noindent
  \begin{tikzpicture}[baseline=(current bounding box.north)]
    \node[anchor=south, align=center,
        minimum width=\leftw, inner sep=1pt]
        at (\leftcx,-\fy + 0pt) {#2};
    \node[anchor=north west, inner sep=0pt,
          minimum width=\fivew, minimum height=\frameh] at (0,-\fy)
      {\includegraphics[width=\fivew,height=\frameh,keepaspectratio]{figure/frame_0001_aug#1.jpg}};
    \node[anchor=north west, inner sep=0pt,
          minimum width=\fivew, minimum height=\frameh] at (\colbx,-\fy)
      {\includegraphics[width=\fivew,height=\frameh,keepaspectratio]{figure/frame_0016_aug#1.jpg}};
    \node[anchor=north west, inner sep=0pt,
          minimum width=\fivew, minimum height=\frameh] at (\colcx,-\fy)
      {\includegraphics[width=\fivew,height=\frameh,keepaspectratio]{figure/frame_0031_aug#1.jpg}};
    \node[anchor=north, inner sep=0pt] at (\graphcx,0)
      {\includegraphics[height=\gh,keepaspectratio]{figure/heatmap_1_frames_aug#1_progress.png}};
  \end{tikzpicture}%
  \par\vspace{\vgap}%
}

\begin{figure}[h]
\centering
\begin{tikzpicture}
  \node[align=center,
        minimum width=\stripw, minimum height=\titleh, inner sep=0pt]
    {\textbf{Reference Demonstration:} Task 4 --- Bimanual Clothes Folding as seen in Figure 9};
\end{tikzpicture}

\vspace{\vgap}

\noindent
\imgbox{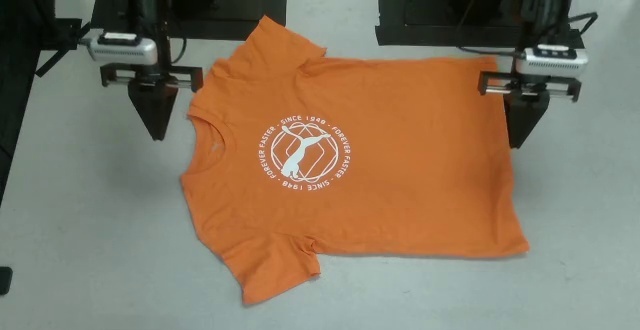}\hspace{\hgap}%
\imgbox{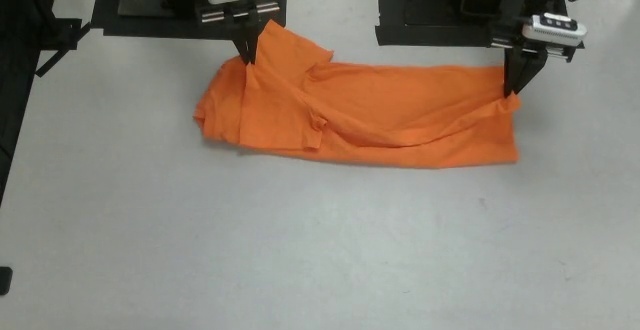}\hspace{\hgap}%
\imgbox{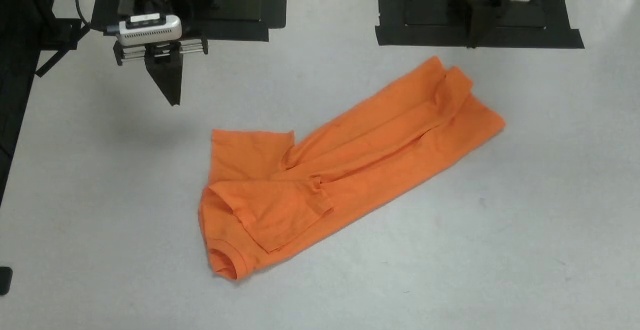}\hspace{\hgap}%
\imgbox{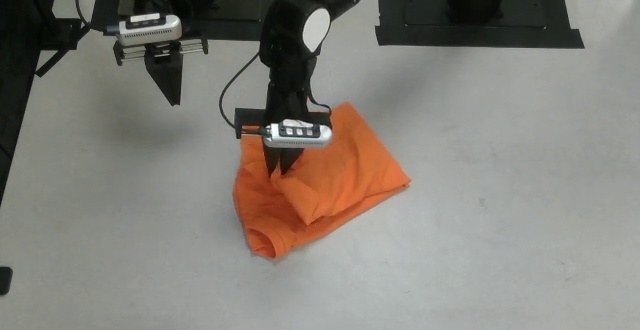}\hspace{\hgap}%
\imgbox{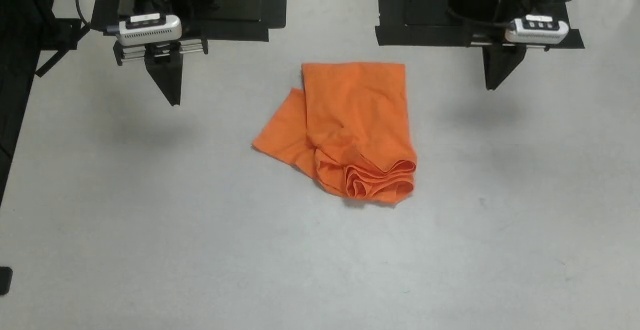}
\noindent
\parbox[t]{\fivew}{\centering\footnotesize Flat orange\\ shirt}\hspace{\hgap}%
\parbox[t]{\fivew}{\centering\footnotesize Bimanual\\ first fold}\hspace{\hgap}%
\parbox[t]{\fivew}{\centering\footnotesize Cooperative\\ $45^{\circ}$ rotation}\hspace{\hgap}%
\parbox[t]{\fivew}{\centering\footnotesize Left pinches\\ near side}\hspace{\hgap}%
\parbox[t]{\fivew}{\centering\footnotesize Final folded\\ shape}%

\vspace{\vgap}

\bigrow{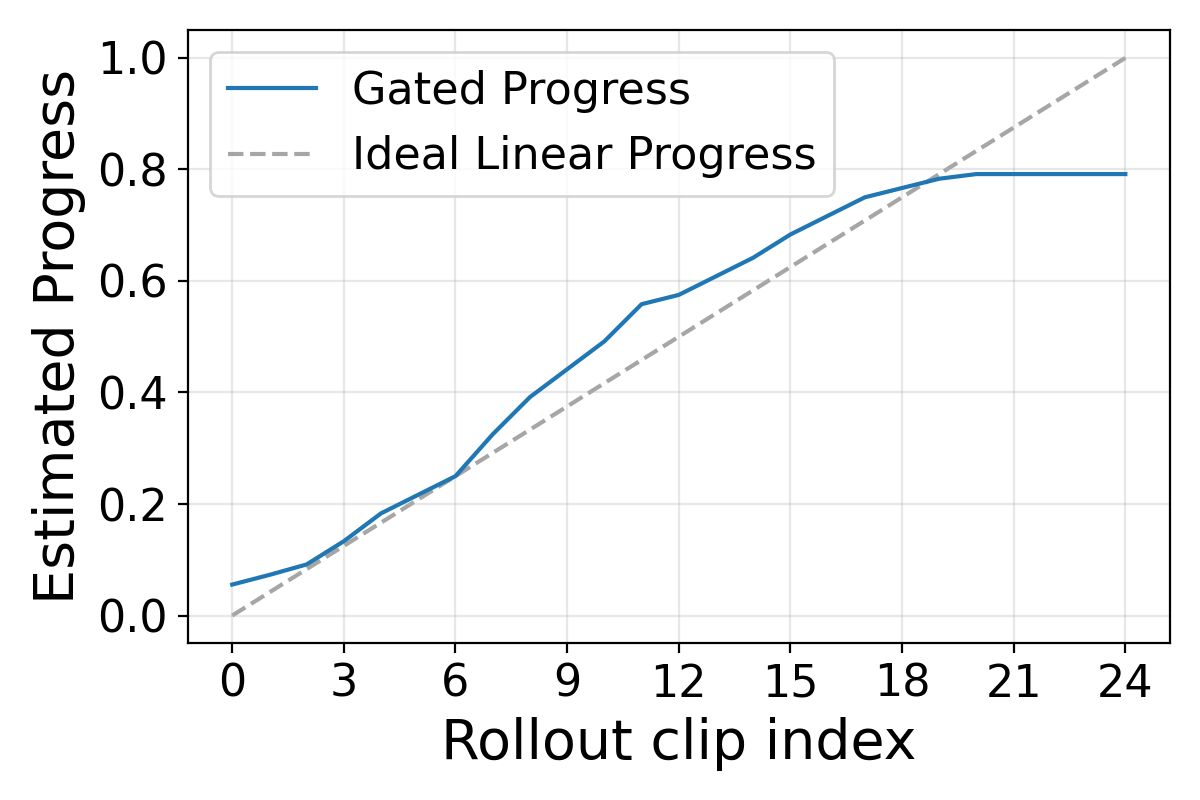}{\textbf{Augment 1 (Glare)}: with blinding white spots}
\bigrow{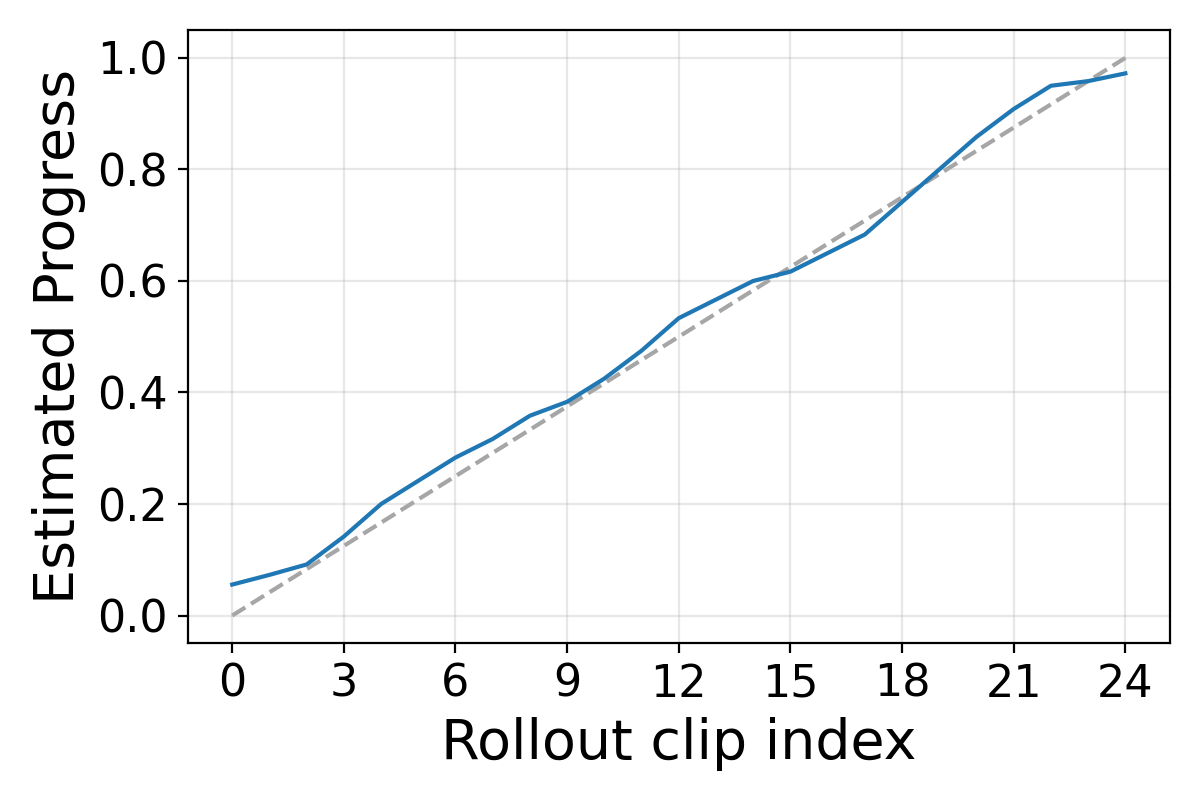}{\textbf{Augment 2 (Color Shift)}: with green filter}
\bigrow{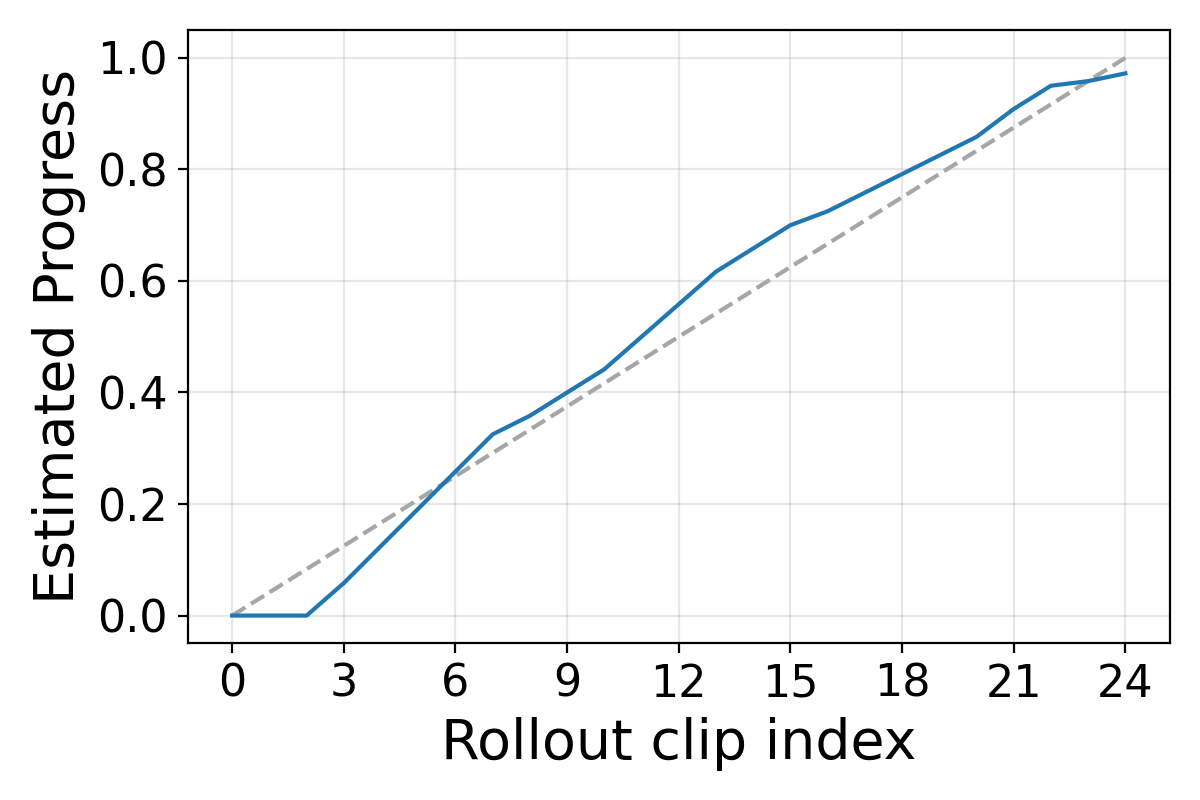}{\textbf{Augment 3 (Occlusion)}: with black box}
\bigrow{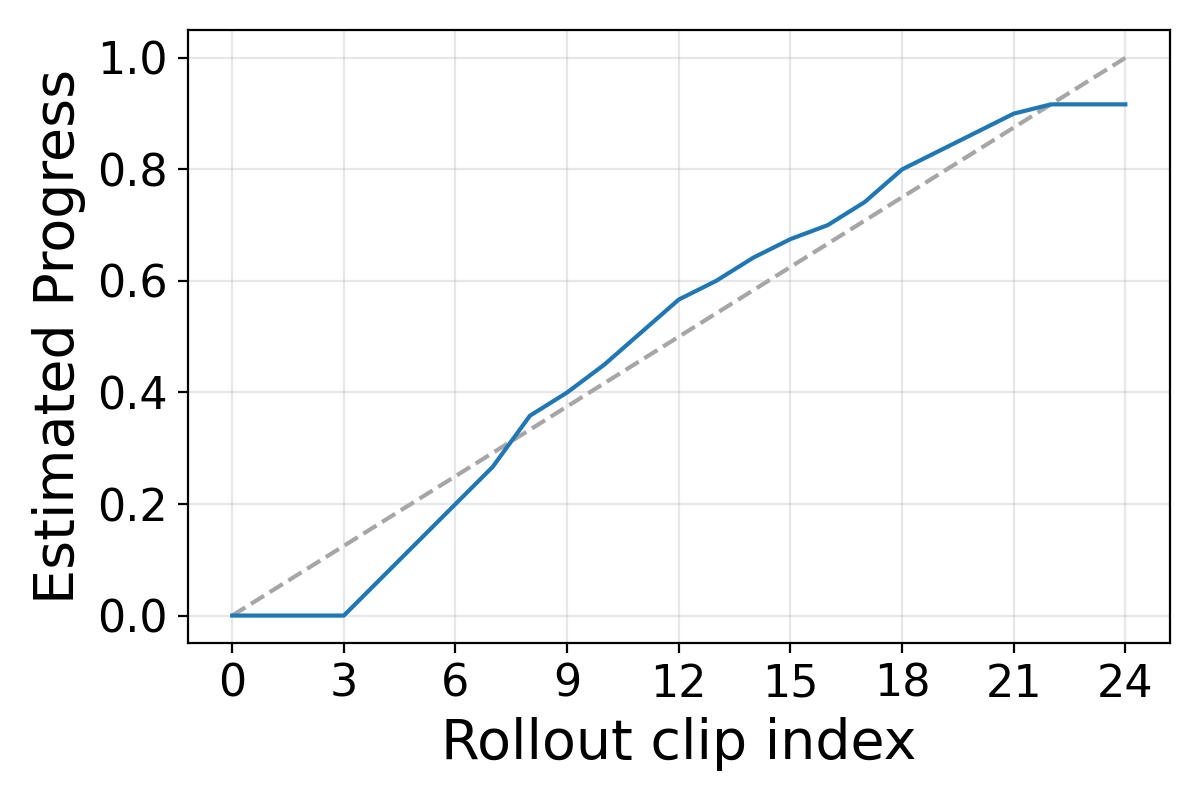}{\textbf{Augment 4 (Viewpoint)}: with perspective pitch and yaw}
\caption{\textbf{RARM Robustness}: Comparison of RARM progress estimates on augmented rollouts compared with a fixed reference demonstration.}
\label{fig:rarm_robustness}
\end{figure}

\subsection{RARM Robustness to Visual Perturbations}
\label{sec: RARM Robustness}
A reward model deployed in the real world must produce stable progress estimates despite the visual variability of uncontrolled environments. To assess this robustness, we conducted a controlled perturbation study using Task 4 from our Real-World Experiments. Starting from a single successful rollout trajectory, we digitally augmentations to its frames and then used RARM to re-estimate the progress of the altered rollout against a different, but unaltered, reference demonstration. 

We evaluated four augmentations, each targeting a distinct real-world failure mode:
\begin{itemize}
    \item \textbf{Augment 1 (Glare):} synthetic white spots and halos that partially occlude the rollout frames, simulating specular highlights and lens flare.
    \item \textbf{Augment 2 (Color shift):} a green tint applied across the frame, simulating changes in ambient lighting and white balance.
    \item \textbf{Augment 3 (Occlusion):} a black box covering part of the image, simulating a foreground obstruction.
    \item \textbf{Augment 4 (Viewpoint):} a perspective warp introducing pitch and yaw, simulating camera misalignment.
\end{itemize}

For every augmentation, RARM continued to recover a coherent progress signal. We visualize this by overlaying the ideal linear progress of the successful rollout against the progress estimated by RARM (Figure~\ref{fig:rarm_robustness}). In all cases the estimated progress closely tracks the ideal linear trend, deviating only by a small margin, indicating that RARM's progress estimates are robust to occlusion, lighting changes, and viewpoint shifts.

\subsection{Inference Speed and GPU Cost Comparisons}
To benchmark the inference speed and CUDA memory usage of our RARM against the baseline reward models, we ran the reward model inference on 125 images composing of 1 rollout trajectory. As seen in Table~\ref{tab:baseline-cost}, our RARM performs in the same band as the rest, being able to process each frame of the 125 in under 10ms. It also uses much less GPU memory compared to VLM-based Reward Models (GVL, Robometer, RoboDopamine). 

\begin{table}[h]
\centering
\caption{Per-frame inference cost and peak memory by reward-model baseline.}
\begin{tabular}{lrr}
\toprule
Baseline & ms/frame & Mem (MB) \\
\midrule
GVL          & $<50$  & 13200 \\
Robometer    &  $<100$ & 13200 \\
RoboDopamine &  $<300$ & 19000 \\
VIP          &  $<10$  & 4200  \\
RoboCLIP     &  $<10$  & 5400  \\
TemporalOT   &  $<10$  & 6600  \\
\textbf{RARM (Ours)}    & $\boldsymbol{<10}$  & \textbf{5100}  \\
\bottomrule
\end{tabular}
\label{tab:baseline-cost}
\end{table}

% \subsection{Additional Ablation}
% Additionally, we conducted another ablation on the network architecture by removing the temporal processor, shown as w/o Temporal alongside the previous ablations. The reward model without its temporal processor performed poorly. This is expected as the temporal processor is the only component in the comparator that gives the clip representation any sense of time. Without it, the model can no longer distinguish forward from reversed motion or early frames from late ones. Since the reward model is used to estimate task progress, losing this directionality makes the per-clip similarities noisier and poorly aligned with true progress, so RL training receives a much weaker, less directional learning signal, seen through the worse ablation results on Figure~\ref{fig:ablation_additional}.

% \begin{figure}[h]
%   \centering
%   \panel{0.48\linewidth}{LIB10: stove + pot}{\includegraphics[width=0.48\linewidth]{figure/abl_lb2.png}}\hfill
%   \panel{0.48\linewidth}{MT: drawer open}{\includegraphics[width=0.48\linewidth]{figure/abl_mw18.png}}
%   \caption{\textbf{Additional Ablation.} Ablation of without temporal processor shown with all previous ablations on one task from LIBERO and MetaWorld.}
%   \label{fig:ablation_additional}
% \end{figure}

\section{Utilized Approaches}
\subsection{Algorithmic Backbone: Drq-v2}

We train our simulation policies from scratch using Drq-v2~\cite{yarats2021mastering}, an off-policy, image-based model-free reinforcement learning algorithm optimized for continuous robotic control. Drq-v2 builds upon the deterministic actor-critic framework to achieve high sample efficiency and stability directly from pixel observations.

The policy training architecture consists of five core components:

\textbf{Visual Regularization}: The environment provides RGB observations, and consecutive frames are stacked to capture short-term temporal context. During optimization, random-shift augmentation is applied to pixel observations, encouraging invariance to small spatial translations and reducing visual overfitting.

\textbf{Random State Initialization in Libero:}
To improve exploration, we initialize training episodes from intermediate states of a task demonstration rather than always resetting to the default initial state. At reset time, the environment restores a non-terminal simulator state. This exposes the policy to diverse progress regions and helps DrQ-v2 learn from later-stage frames that are rarely reached under random exploration.

% \textbf{Actor-Critic Architecture:} A shared convolutional encoder extracts visual features for both the policy and value networks. These features are projected through a compact trunk representation. The deterministic actor maps this representation to continuous actions, with exploration controlled by Gaussian noise following a decay schedule. Full architectural details are provided in Table~\ref{tab:drqv2_architecture}, and optimization hyperparameters are listed in Table~\ref{tab:hp-drqv2-opt}.

\textbf{Actor-Critic Architecture:}
The actor and critics share a convolutional encoder, followed by separate
actor and critic projection trunks. Each trunk maps the visual features
to a 50-dimensional representation using a linear layer, layer normalization, and a tanh activation. The two Q-functions share the critic trunk. The actor and each Q-function use two hidden layers of 1024 units with ReLU activations. The actor outputs continuous actions, with exploration provided by truncated Gaussian noise following a decay schedule. Full architectural details are provided in
Table~\ref{tab:drqv2_architecture}, and optimization hyperparameters
are listed in Table~\ref{tab:hp-drqv2-opt}.

\textbf{Value Optimization:} DrQ-v2 uses twin critics together with EMA target networks to stabilize value learning. Each critic estimates the expected return from the encoded observation and action, and is trained with multi-step temporal-difference targets from an off-policy replay buffer:

\begin{equation}
    \mathcal{L}(\theta_i) = \mathbb{E}_{(\mathbf{s}, \mathbf{a}, r, \mathbf{s}') \sim \mathcal{D}} \left[ \left( Q_{\theta_i}(\mathbf{s}, \mathbf{a}) - \left( \sum_{k=0}^{2} \gamma^k r_{t+k} + \gamma^3 \min_{j=1,2} Q_{\bar{\theta}_j}(\mathbf{s}_{t+3}, \mathbf{a}_{t+3}) \right) \right)^2 \right]
\end{equation}

\textbf{RARM Reward Integration.}
RARM scores buffered rollout clips while DrQ-v2 acts on
low-resolution stacked observations. Clip rewards are assigned
to their middle sampled frames, with positive rewards extended
to the two preceding sampled positions and held constant between
samples. The resulting dense signal is scaled by the reward-model
weight and combined with a task-success reward of $5$ and a
per-step cost of $0.5$ to form the reward $r_t$ used in the
DrQ-v2 Bellman update.

\subsection{pi0}
\textbf{VLA backbone.}
We use $\pi_0$ (Pi-Zero) as the base VLA policy on AIRBOT Play. 
The model predicts smooth continuous action trajectories from RGB observations and language instructions using a flow-matching-based action generator. 
In our framework, $\pi_0$ provides the initial policy, while DSRL further improves it using either sparse binary rewards or our RARM dense reward.

\textbf{Deployment details.}
The policy predicts $50$-step action chunks on a $50$\,Hz temporal grid and replans every $25$ executed steps. 
The predicted actions are sent to the AIRBOT Play control interface, which runs at approximately $20$\,Hz in our real-robot setup. 
The policy consumes $128\times128$ RGB observations, while camera streams are ordered as environment, left-wrist, and right-wrist views. 

% Inference requires at least $8$\,GB GPU memory; LoRA fine-tuning is feasible on approximately $24$\,GB GPUs.

\subsection{DSRL: Diffusion Steering RL for Real-Robot Adaptation}
\label{app:dsrl}

For real-robot experiments we use Diffusion Steering RL (DSRL)~\citep{wagenmaker2025steering} to adapt a frozen $\pi_0$ diffusion policy~\citep{black2024pi_0} to each task. A pixel-based SAC agent learns to predict the latent noise that seeds $\pi_0$'s diffusion process; $\pi_0$ itself is never updated. RARM produces one reward per $\pi_0$ query step from the captured rollout frames, replacing the default $-1$ rewards in default DSRL implementation.

The SAC actor and twin Q-critics share a small CNN encoder over $128{\times}128$ multi-camera RGB inputs and 3-layer MLPs with hidden size $1024$. Batch size $256$, $\gamma{=}0.99$, learning rates $10^{-4}$ (actor) and $3\!\times\!10^{-4}$ (critic, temperature), target entropy $0$. The first $10$ rollouts use random noise to fill in the buffer. Comparatively, RARM uses $384\times384$ images for higher-fidelity visual comparison. Task-specific settings:

\begin{center}
\small
\begin{tabular}{lccc}
\toprule
Task & \texttt{max\_timesteps} & Query steps & Episode Length (seconds) \\
\midrule
fold\_clothes & 1000 & 40 & 50 \\
open\_drawer  & 175  & 7  & 8.75 \\
pick\_eraser  & 200  & 8  & 10 \\
handover      & 300  & 12 & 15 \\
\bottomrule
\end{tabular}
\end{center}

Each task uses one reference demonstration (frames of a successful rollout from initial $\pi_0$); no progress labels or task-specific reward engineering are required. At the end of every rollout the operator records success ($1$) or failure ($0$).

\end{document}